\pdfoutput=1
\documentclass[10pt,twocolumn,letterpaper]{article}

\usepackage[expansion=false]{microtype}
\usepackage[pagenumbers]{wacv}
\usepackage[accsupp]{axessibility}
\usepackage{colortbl}
\usepackage{multirow}
\usepackage{cuted}
\usepackage{afterpage}
\usepackage{makecell}

\definecolor{colFirst}{RGB}{255,130,110}
\definecolor{colSecond}{RGB}{255,196,90}
\definecolor{colThird}{RGB}{255,238,158}
\newcommand{\hlbox}[3]{{\setlength{\fboxsep}{1pt}\colorbox{#1}{\strut\hspace{#2}#3\hspace{#2}}}}

\newcommand{\hFS}[1]{\hlbox{colFirst}{1.5pt}{\textbf{#1}}}
\newcommand{\hSS}[1]{\hlbox{colSecond}{1.5pt}{\underline{#1}}}
\newcommand{\hTS}[1]{\hlbox{colThird}{1.5pt}{\textit{#1}}}

\newcommand{\hFM}[1]{\hlbox{colFirst}{5.0pt}{\textbf{#1}}}
\newcommand{\hSM}[1]{\hlbox{colSecond}{5.0pt}{\underline{#1}}}
\newcommand{\hTM}[1]{\hlbox{colThird}{5.0pt}{\textit{#1}}}

\newcommand{\hFL}[1]{\hlbox{colFirst}{9.0pt}{\textbf{#1}}}
\newcommand{\hSL}[1]{\hlbox{colSecond}{9.0pt}{\underline{#1}}}
\newcommand{\hTL}[1]{\hlbox{colThird}{9.0pt}{\textit{#1}}}

\newcommand{\hF}[1]{\hlbox{colFirst}{4.0pt}{\textbf{#1}}}
\newcommand{\hS}[1]{\hlbox{colSecond}{4.0pt}{\underline{#1}}}
\newcommand{\hT}[1]{\hlbox{colThird}{4.0pt}{\textit{#1}}}

\newlength{\cnshift}
\newcommand{\ciii}[1]{#1\hspace{2\cnshift}}
\newcommand{\cii}[1]{\hspace{\dimexpr\cnshift+0.25em\relax}#1\hspace{\dimexpr\cnshift+0.25em\relax}}
\newcommand{\ci}[1]{\hspace{\dimexpr2\cnshift+0.5em\relax}#1\hspace{0.5em}}

\definecolor{wacvblue}{rgb}{0.21,0.49,0.74}
\usepackage[pagebackref,breaklinks,colorlinks,allcolors=wacvblue]{hyperref}
\AtBeginDocument{\crefname{appendix}{Sec.}{Secs.}}
\makeatletter
\renewcommand\LenToUnit[1]{\strip@pt\dimexpr#1\relax}
\makeatother

\title{Mamba2D: A Natively Multi-Dimensional State-Space Model for Vision Tasks}

\author{
Enis Baty$^{1,}$\thanks{Equal contribution.}\hspace{0.9em}%
Alejandro Hernández Díaz$^{1,*}$\hspace{0.9em}%
Rebecca Davidson$^{2}$\hspace{0.9em}%
Chris Bridges$^{1}$\hspace{0.9em}%
Simon Hadfield$^{1}$\\
$^{1}$University of Surrey \qquad
$^{2}$Surrey Satellite Technology Limited\\
{\tt\small \{e.baty, a.hernandez, c.p.bridges, s.hadfield\}@surrey.ac.uk \qquad bdavidson@sstl.co.uk}
}

\begin{document}
\maketitle

\setlength{\stripsep}{-1.5cm}
\begin{strip}
    \vspace{-1.5em}
    \centering
    \begin{minipage}[c]{0.54\linewidth}
        \centering
        \begin{minipage}[b][0.28\textheight][b]{\linewidth}
            \centering
            \includegraphics[width=\linewidth]{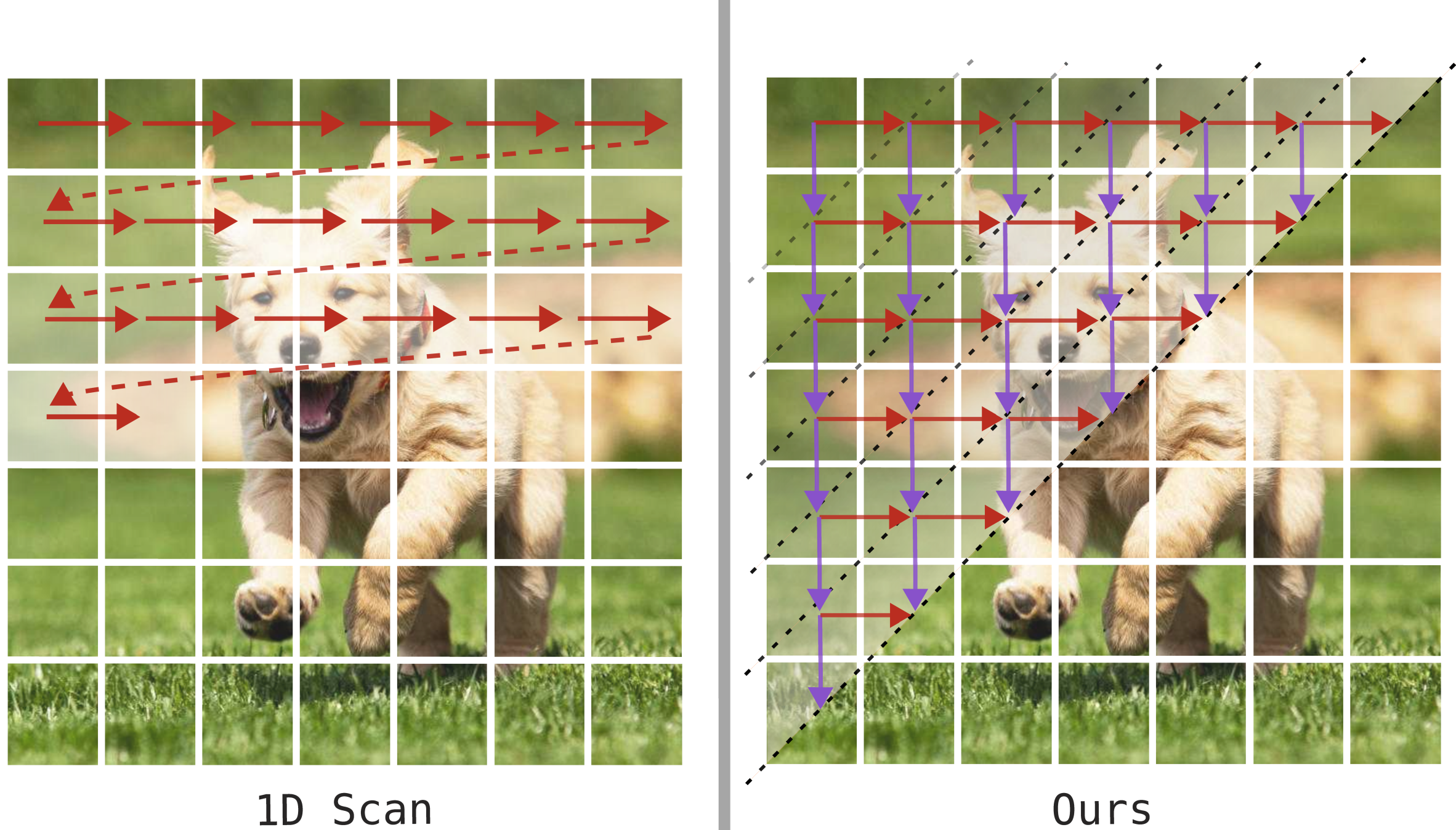}
        \end{minipage}
        \vspace{0.35em}
        \vspace{0.40em}
    \end{minipage}
    \begin{minipage}[c]{0.38\linewidth}
        \centering
        \begin{minipage}[b][0.28\textheight][b]{\linewidth}
            \centering
            \includegraphics[width=\linewidth]{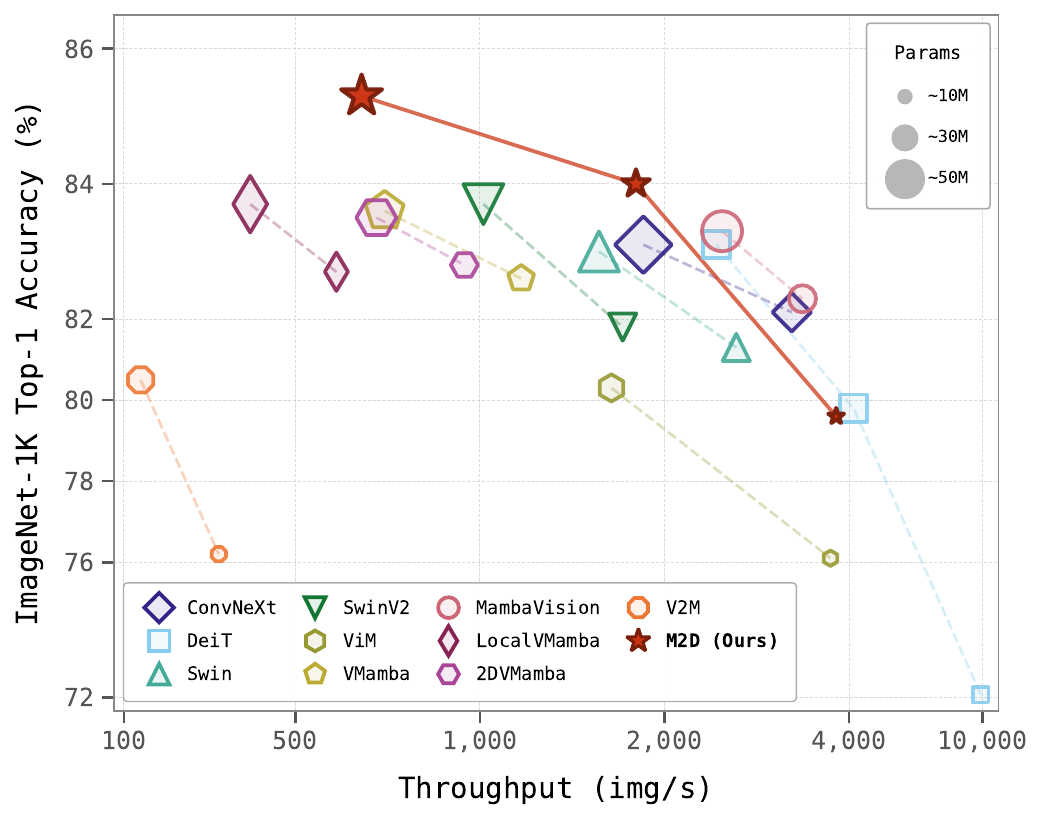}
        \end{minipage}
        \vspace{0.18em}
    \end{minipage}
    \vspace{0.2em}
    \captionof{figure}{\textbf{(a)} Comparison of traditional rasterised scanning (left) vs.\ our native 2D wavefront scan (right); diagonals indicate parallel computation steps.
             \textbf{(b)} M2D (stars) establishes a new accuracy-throughput-params frontier for SSM vision models.}
    \label{fig:teaser}
    \vspace{\dimexpr1.7cm+1.0em\relax}
\end{strip}
\begingroup\setlength{\parskip}{0pt}\def\endabstract{\par\vspace{12pt}}
\begin{abstract}
State-Space Models (SSMs) have emerged as an efficient alternative to transformers,
yet existing visual SSMs retain deeply ingrained biases from their origins in natural
language processing.
In this paper, we address
these limitations by introducing M2D-SSM, 
a ground-up re-derivation of selective state-space techniques for
multidimensional data. Unlike prior works that apply 1D SSMs directly to images through
arbitrary rasterised scanning, our M2D-SSM employs a single 2D scan
that factors in both spatial dimensions natively. We pair M2D-SSM blocks at high resolution with
vanilla attention at low resolution in a hybrid backbone, allowing each mechanism to operate in its
optimal regime without restriction.
Across dense prediction tasks, where the choice of token mixer is most consequential, Mamba2D achieves 52.2 box AP on MS-COCO object detection (3× schedule) and 51.7 mIoU on ADE20K segmentation. On ImageNet-1K classification, M2D-T reaches 84.0\% top-1 with only 27M parameters and M2D-S 85.3\%, matching or exceeding all prior SSM-based vision models at comparable size. Code and models are available at \url{https://github.com/cocoalex00/Mamba2D}.
\end{abstract}
\endgroup

\section{Introduction}
\label{sec:intro}
\vspace{5pt}%

Since the advent of the Vision Transformer (ViT) \cite{dosovitskiyImageWorth16x162021a}, countless works have
explored transformers across visual tasks \cite{yuMetaFormerActuallyWhat2022,daiCoAtNetMarryingConvolution2021,guoCMTConvolutionalNeural2022,kagAsCANAsymmetricConvolutionAttention2024,liuSwinTransformerHierarchical2021,liuSwinTransformerV22022,bochkovskiiDepthProSharp2024}, reshaping how image data is processed and interpreted.
Yet attention scales quadratically with image size, an overhead that becomes prohibitive for
high-resolution and dense tasks such as detection, segmentation, remote sensing and medical imaging. In
response, language modelling has turned to state-space models (SSMs), which scale linearly in input
length. In particular, Mamba \cite{guMambaLinearTimeSequence2023} introduced a selectivity mechanism that aggregates
global context dynamically, analogously to attention.

However, adopting SSMs for vision exposes biases inherited from language modelling: (1) SSMs were
designed for 1D signals and cannot be applied to
multidimensional data without reshaping; (2) their causal structure,
suited to language, is at odds with the predominantly non-causal
requirements of vision.
Previous works partially address these limitations by unrolling 2D
images into multiple 1D sequences, scanned individually and fused to
capture spatial relationships.
Unfortunately, the recurrent memory of SSMs
provides exponentially decaying recall with distance.
This makes them sensitive to data locality, which no combination of 2D-to-1D mappings can preserve.

In contrast, we fundamentally re-design the selective State-Space paradigm to
produce the M2D-SSM: a natively 2D block that preserves the
selectivity of Mamba while propagating information along both spatial
axes in a single joint recurrence. Spatial neighbours thus remain one
recurrence step apart, whereas prior raster flattening approaches separate vertical
neighbours by an entire image row. These 2D dependencies are incompatible with
parallel scan approaches, requiring a custom CUDA wavefront-scan
kernel (\cref{sec:cuda-wf-impl}). We further construct a hybrid architecture
pairing M2D-SSM blocks with vanilla attention: M2D-SSM handles
high-resolution stages where its linear scaling excels, while
vanilla attention is applied at low-resolution stages, where spatial
cost is minimal.

To the best of our knowledge, no prior work derives a native 2D selective SSM
for vision without workarounds that sacrifice 2D spatial structure.
Our primary contributions are:
(1) A natively 2D SSM formulation that faithfully operates on both
    spatial dimensions without 1D flattening.
(2) A custom fused CUDA wavefront-scan kernel enabling parallelisable
    training of the 2D recurrence.
(3) A hybrid architecture that synergistically combines M2D-SSMs at
    high resolution with vanilla attention at low resolution, with neither mechanism restricted: no
    windowed or factored attention, and no decomposition of the 2D scan into 1D passes.
(4) State-of-the-art results among SSM-based models on  ImageNet-1K
    classification, MS-COCO detection, and ADE20K segmentation.

\section{Related Work}
\label{sec:related}

The Vision Transformer (ViT) \cite{dosovitskiyImageWorth16x162021a}
demonstrated that patch-based transformers can match or surpass CNNs
on large-scale image classification, with DeiT \cite{touvronTrainingDataefficientImage2021a}
improving data efficiency via distillation. However, the quadratic
cost of global self-attention limits scalability to high-resolution
inputs. This motivated hierarchical variants such as Swin \cite{liuSwinTransformerHierarchical2021,liuSwinTransformerV22022} and PVT \cite{wangPyramidVisionTransformer2021}, which use local windows or progressive spatial reduction. Modernised CNNs such
as ConvNeXt \cite{liuConvNet2020s2022} remain competitive, and various
hybrid architectures have been proposed
\cite{daiCoAtNetMarryingConvolution2021,guoCMTConvolutionalNeural2022,kagAsCANAsymmetricConvolutionAttention2024}
to leverage the complementary strengths of both paradigms.

State-space models (SSMs) offer a compelling alternative for sequence modelling, driven by linear
scaling with sequence length. S4 \cite{guEfficientlyModelingLong2022} established that SSMs can model
long-range dependencies through efficient parametrisation and HiPPO initialisation
\cite{guHiPPORecurrentMemory2020}. Mamba \cite{guMambaLinearTimeSequence2023} then introduced an
input-dependent selective scan (S6), letting SSMs aggregate global context dynamically, akin to
attention, a connection that Mamba-2 \cite{daoTransformersAreSSMs2024} formalised as State
Space Duality (SSD). These advances establish SSMs as powerful sequence modellers, but their
formulations remain inherently one-dimensional.

Vision adaptations of Mamba therefore flatten 2D images into 1D sequences. Vision Mamba
\cite{zhuVisionMambaEfficient2024} applies bidirectional 1D scanning within an isotropic architecture,
while VMamba \cite{liuVMambaVisualState2024} processes four directional scans through a cross-scan
module in a hierarchical design. Variations include EfficientVMamba
\cite{peiEfficientVMambaAtrousSelective2024} with atrous scanning, PlainMamba \cite{plainmamba} with
non-hierarchical structures, LocalMamba \cite{localmamba} restricting scans to local windows,
GroupMamba \cite{shakerGroupMambaEfficient2025} with distillation-trained grouped scanning, and
EfficientViM \cite{leeEfficientViMHidden2025} with hidden-state SSD mixing.

All of these flattening approaches share a fundamental limitation: they destroy the native 2D spatial
relationships between neighbouring patches. Because an SSM's recurrent memory decays exponentially with
sequence distance, patches that are spatially adjacent but sequentially distant receive
disproportionately weak coupling. Multiple scan directions partially compensate, but add computational
redundancy without resolving the underlying problem.

A smaller body of work extends SSMs directly to two spatial dimensions. S4ND
\cite{nguyenS4NDModelingImages2022} generalises S4 to multiple dimensions and 2D-SSM \cite{baron2DimensionalStateSpace2024} adopts the classical Roesser 2D state-space model, but both are linear
time-invariant (LTI) and hence lack input-dependent selectivity. V2M
\cite{wangV2MVisual2Dimensional2024} and 2DMamba \cite{zhang2DMambaEfficientState2024} propose 2D selective scans, yet
both decompose the operation into sequential horizontal and vertical 1D passes. VSSD
\cite{shiVSSDVisionMamba2024} instead removes the causal mask from SSD, collapsing the recurrence into
a global weighted sum over all tokens and abandoning sequential scanning for a mechanism closer to
linear attention. Ultimately, none of these derive a truly native selective 2D SSM from first principles: they
either reduce to compositions of 1D scans, or dispense with recurrent scanning.
Outside the SSM family, joint 2D recurrences long predate SSMs: MDLSTM \cite{gravesMultidimensionalRecurrentNeural2007}
and Grid LSTM \cite{kalchbrennerGridLongShortTerm2016} propagate gated hidden states along both spatial axes, and
PixelRNN \cite{oordPixelRecurrentNeural2016} evaluates such a recurrence one diagonal at a time. More recently,
GSPN \cite{wangParallelSequenceModeling2025} applies row-by-row linear propagation in four directional passes.
M2D-SSM brings the joint 2D recurrence to selective SSMs.

Orthogonal to the scan formulation, hybrid designs combine SSM and attention layers, and can
outperform either mechanism alone. MambaVision
\cite{hatamizadehMambaVisionHybridMambaTransformer2024} interleaves Mamba blocks and self-attention
\textit{within} each stage; we instead separate the two \textit{across} stages, assigning each
mechanism to the resolution regime that suits it.

\section{Preliminaries}

State-space models describe a 1D mapping from the continuous signal $x(t): \mathbb{R} \rightarrow \mathbb{R}$ to $y(t): \mathbb{R} \rightarrow \mathbb{R}$ through an N-D hidden state $h(t): \mathbb{R} \rightarrow  \mathbb{R}^N$. Such a model is parametrised by two projection matrices to and from the latent space ($\mathbf{B} \in \mathbb{R}^{N\times1}$, $\mathbf{C}\in \mathbb{R}^{1\times N}$) and an evolution matrix $\mathbf{A} \in \mathbb{R}^{N\times N}$. The full system is then formulated as linear ordinary differential equations (ODEs)  
\begin{equation}
\begin{aligned}
h^{\prime}(t) & =\mathbf{A} h(t)+\mathbf{B} x(t), \\
y(t) & =\mathbf{C} h(t) + \mathbf{D} x(t),
\end{aligned}\label{eq:11}
\end{equation}
where $h'(t)$ is the derivative of the hidden state at time $t$, and $\mathbf{D}$ is a skip parameter that allows the input to directly influence the output.
As observations exist only at discrete intervals, we discretise via Zero Order Hold (ZOH) \cite{guEfficientlyModelingLong2022}:
\begin{equation}
    \mathbf{\overline{A}}=\exp (\Delta \mathbf{A}) \quad
    \mathbf{\overline{B}}=(\Delta \mathbf{A})^{-1}(\exp (\Delta \mathbf{A})-\mathbf{I}) \cdot \Delta \mathbf{B},
\end{equation}
where $\Delta$ is a learnable step size controlling the influence of each input token on the hidden state.
This yields a discrete-time recurrence,
\begin{equation}
    \begin{aligned}
    & h_t=\mathbf{\overline{A}} h_{t-1}+\mathbf{\overline{B}} x_t, \\
    & y_t=\mathbf{C} h_t + \mathbf{D} x_t,
    \end{aligned}
    \label{eq:recurrence}
\end{equation}
which scales linearly in complexity with sequence length.

S6/Mamba adds an input-dependent parametrisation for context-based reasoning, learning time-varying functions $\mathbf{B}_t=S_{\mathbf{B}}(x_t)$, $\mathbf{C}_t=S_{\mathbf{C}}(x_t)$, $\Delta_t=\tau_{\Delta}\!\left(\Delta+S_{\Delta}(x_t)\right)$ that transform the input elements into said parameters.

\begin{figure*}[t]
    \centering
    \includegraphics[width=0.85\textwidth]{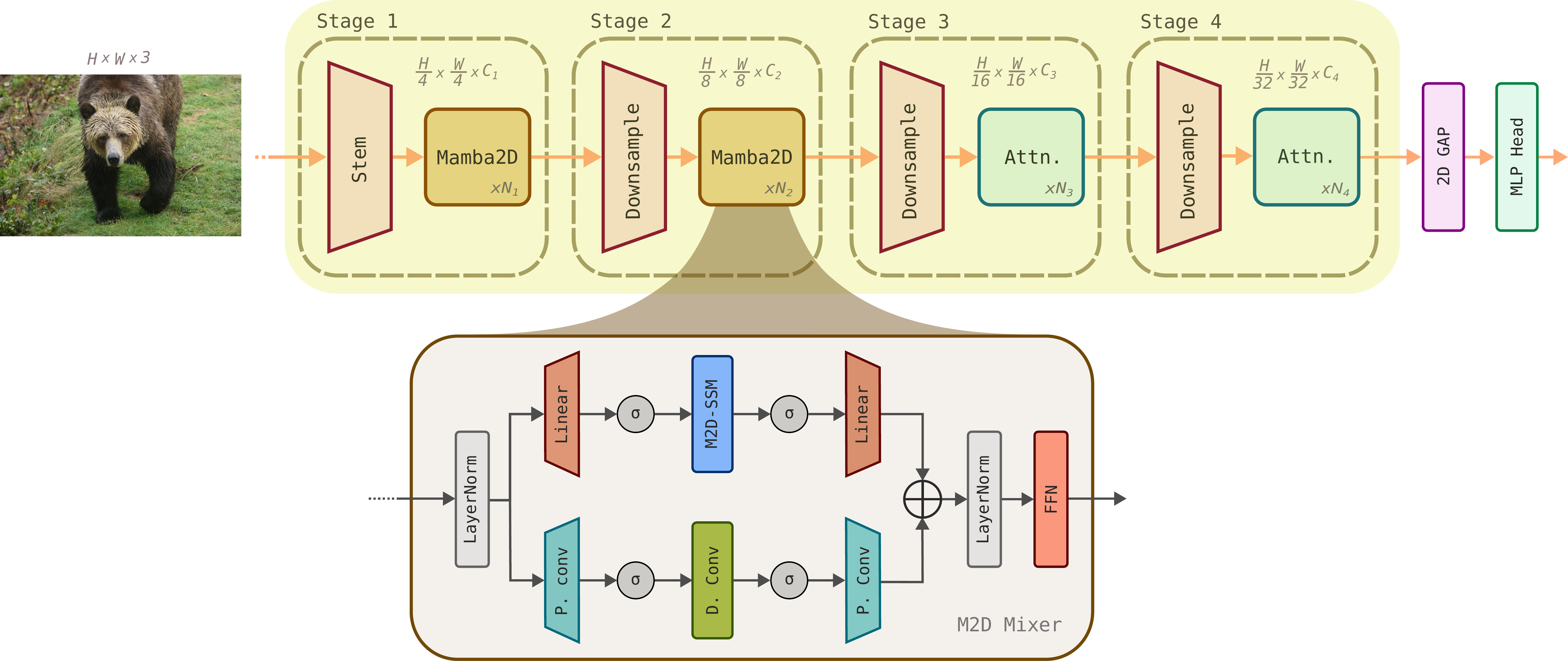}
    \caption{Mamba2D model architecture. A convolutional stem performs patch
             embedding, followed by four stages of feature
             extraction. Each stage consists of $N_{1 \ldots 4}$ blocks, comprising
             a token mixer and an FFN. The Mamba2D token mixer uses two
             parallel branches: our native 2D SSM path and a local processing path,
             whose outputs are combined before the FFN.
             }
    \label{fig:2DM-model-arch}
\end{figure*}

\section{Methodology}\label{sec:method}

Our reformulation of Mamba for 2D inputs partly follows S4ND, initialising
two independent sets of axis-specific parameters
(\ie $\mathbf{A}_{t},\mathbf{B}_{t},\Delta_{t}$ and $\mathbf{A}_{z},\mathbf{B}_{z},\Delta_{z}$),
with output parameters $\mathbf{C}$ and $\mathbf{D}$ shared across both axes.
However, we contrast with S4ND in two key respects. First, S4ND constructs its 2D output by taking the outer product of two independent 1D SSM transfer functions, which constrains the 2D hidden state to a rank-1 product of separate row and column states. Our formulation instead uses additive joint recurrence, where $h(t,z)$ is simultaneously influenced by both $h(t-1,z)$ and $h(t,z-1)$, enabling richer cross-axis interactions that cannot be expressed as a product of independent 1D states. Furthermore, S4ND is restricted to LTI parameters, whereas our formulation inherits the input-dependent selectivity of Mamba, allowing the model to dynamically modulate information flow based on content.

Formally, we extend \cref{eq:11} to a 2D domain by introducing two independent partial derivatives ($h'_t$ and $h'_z$):
\begin{gather}
    h_t'(t,z) = \mathbf{A}_th(t,z) + \mathbf{B}_tx(t,z), \label{eqn:ode-2d-hp} \\
    h_z'(t,z) = \mathbf{A}_zh(t,z) + \mathbf{B}_zx(t,z). \label{eqn:ode-2d-hp-b}
\end{gather}

We form the discrete recurrence by applying ZOH discretisation along each spatial axis. Since each partial derivative (\cref{eqn:ode-2d-hp,eqn:ode-2d-hp-b}) governs dynamics along a single axis, the standard 1D ZOH derivation \cite{guEfficientlyModelingLong2022} applies directly to each, giving
\begin{gather}
    h(t + \Delta_t, z) = \overline{\mathbf{A}}_t h(t,z) + \overline{\mathbf{B}}_t x(t,z) \label{eqn:2d-disc-t}, \\
    h(t, z + \Delta_z) = \overline{\mathbf{A}}_z h(t,z) + \overline{\mathbf{B}}_z x(t,z) \label{eqn:2d-disc-z},
\end{gather}
where $\overline{\mathbf{A}}_{t} = \exp(\Delta_{t}\mathbf{A}_{t})$ and
$\overline{\mathbf{B}}_{t} = \Delta_{t}\mathbf{B}_{t}$ (and similarly for $z$), following the simplified ZOH convention of \cite{guMambaLinearTimeSequence2023}.

This formulation yields two independent recurrences for the same hidden state (from above via
\cref{eqn:2d-disc-t} and from the left via \cref{eqn:2d-disc-z}), whose relative contributions
are modulated by the input-dependent selective parameters $\overline{\mathbf{A}}_{t,z}$ and
$\overline{\mathbf{B}}_{t,z}$. 
Under the standard Mamba parametrisation $\mathbf{A}$ is diagonal with negative entries and
$\Delta > 0$, thus $\overline{\mathbf{A}}_{t,z}$ has entries in $(0,1]$. Combining the two incoming contributions with a fixed factor of
$\frac{1}{2}$ therefore bounds the per-step gain 
by $\max\|\overline{\mathbf{A}}\| \leq 1$,
preventing unbounded growth of the hidden state.
Supplementary \cref{sec:supp-alpha} contains a more detailed analysis of this
combination factor.
This results in our 2D SSM equations:
\begin{gather} \label{eqn:final-2D-SSM}
    h(t,z) \!=\! \frac{1}{2} \!\left(\!
    \begin{bmatrix}
        \overline{\mathbf{A}}_t \\
        \overline{\mathbf{A}}_z \\
    \end{bmatrix}^{\!\!\top\!\!}
    \begin{bmatrix}
        h(t-\Delta_t,z) \\
        h(t,z-\Delta_z)
    \end{bmatrix}
    \!\!+\!\!
    \begin{bmatrix}
        x(t,z) \\
        x(t,z) \\
    \end{bmatrix}^{\!\!\top\!\!}
    \begin{bmatrix}
        \overline{\mathbf{B}}_t \\
        \overline{\mathbf{B}}_z \\
    \end{bmatrix}
    \!\right)\! \\
    y(t,z) =\mathbf{C}h(t,z) +\mathbf{D}x(t,z).
\end{gather}

The influence of each 2D input element on each 2D output element is thus determined by their
2D offset, rather than the distance along any arbitrary 1D flattening of the data, a property
we measure directly in supplementary \cref{sec:supp-influence}.

To address the causal bias of SSMs, we alternate scan direction (top-left-to-bottom-right
vs.\ bottom-right-to-top-left) between subsequent blocks \cite{zhuVisionMambaEfficient2024}, capturing
bidirectional spatial context without duplicating computation within a block (ablated in \cref{sec:abl-scan-dir}).

\subsection{CUDA 2D Wavefront Scan} \label{sec:cuda-wf-impl}
The associative scan of 1D Mamba cannot be directly extended to parallelise this 2D recurrence,
because
cumulative multiplication of $\overline{\mathbf{A}}_{t}$ and $\overline{\mathbf{A}}_{z}$ creates
an intractable number of cross-diagonal paths. We therefore implement wavefront parallelism
(\cref{fig:teaser}(a)), computing all elements along a diagonal in parallel via a custom CUDA kernel
(\cref{sec:cuda-wf-impl}), while subsequent diagonals are processed sequentially.
We show empirically that this wavefront scan yields both
throughput and accuracy
gains (\cref{sec:scan-op-comparison}, \cref{sec:abl-stage-mixer}).

More formally, within our scan each hidden state $h(t,z)$ depends only on its top $h(t-1, z)$ and left $h(t, z-1)$ neighbours (see \cref{fig:teaser}(a)).
As such, all states along a diagonal (where $t+z=d$) are independent of each other and can be computed
in parallel.

We dispatch diagonal kernels sequentially from $d=0$ to $H+W-2$, where $H$ and $W$ denote the spatial
dimensions of the input. Each kernel computes \cref{eqn:final-2D-SSM} for all elements within
the current diagonal and writes the result to the output tensor $h$. This dispatch order ensures that
the necessary dependencies from the previous wavefront are already committed to global memory before
the next kernel begins execution.

Following \cite{guMambaLinearTimeSequence2023}, scan operators
are memory-bound due to the low arithmetic intensity of the
element-wise recurrence. To mitigate this, our CUDA kernels
fuse the discretisation step: rather than materialising
$\overline{\mathbf{A}}_{t,z}$ and $\overline{\mathbf{B}}_{t,z}$
in global memory, we compute $\exp(\Delta \cdot A)$ and
$\Delta \cdot B \cdot x$ on-the-fly in registers from the
projected continuous-time parameters. We additionally avoid
storing the hidden states $\mathbf{h} \in \mathbb{R}^{B \times
H \times W \times E \times N}$ during training, instead
recomputing them during backpropagation. We quantify the
impact of both strategies in supplementary \cref{sec:supp-recomp}.

\begin{table}
\centering

\caption{Architecture configurations of Mamba2D model variants. \textit{Note: DP Rate refers to Drop Path rate (stochastic depth).}}
\label{tab:model-variants}
\resizebox{1.0\linewidth}{!}{%
\footnotesize{\setlength{\tabcolsep}{4pt}
\begin{tabular}{c|c|c|c|c|c|c}
\toprule
Model & Blocks & Channels & DP Rate & Dropout & \#Param. & FLOPs \\
\midrule
M2D-N & [3, 3, 9, 3]   & [32, 64, 160, 256]   & 0.000   & 0.00  & \phantom{0}7.2M  & \phantom{0}1.35G \\
M2D-T & [3, 3, 9, 3]   & [64, 128, 320, 512]  & 0.175 & 0.15 & 26.5M & \phantom{0}4.66G \\
M2D-S & [3, 12, 14, 3] & [96, 192, 384, 576]  & 0.400   & 0.40  & 50.3M & 13.74G \\
\bottomrule
\end{tabular}}}
\end{table}
\subsection{Mamba2D Model}
\begin{table}[!bt]
\centering
\caption{Top-1 accuracy on ImageNet-1K (FLOPs at $224^2$), separated into attention-based and 1D-SSM methods, followed by convolutional, convolution-attention hybrid and 2D-SSM methods, including our M2D. Results sorted by parameter count within each group. For each model size we highlight the \hFS{1st}\,/\,\hSS{2nd}\,/\,\hTS{3rd} best; highlight width shows model scale.}
\label{tab:class-res}
\resizebox{0.8\linewidth}{!}{%
\footnotesize\setlength{\tabcolsep}{5pt}
\renewcommand{\arraystretch}{1.0}%
\begin{tabular}[t]{l|c|c|c}
\toprule
\textbf{Model} & \textbf{\#Param.} & \textbf{FLOPs} & \textbf{Top-1 (\%)} \\
\midrule
\multicolumn{4}{l}{\textit{Attention}} \\
\midrule
DeiT-S   \cite{touvronTrainingDataefficientImage2021a}     & 22 M  & 4.6 G & 79.8 \\
Swin-T   \cite{liuSwinTransformerHierarchical2021}         & 29 M  & 4.5 G & 81.3 \\
Swin-S   \cite{liuSwinTransformerHierarchical2021}         & 50 M  & 8.7 G & 83.2 \\
DeiT-B   \cite{touvronTrainingDataefficientImage2021a}     & 86 M  & 17.5 G & 81.8 \\
Swin-B   \cite{liuSwinTransformerHierarchical2021}         & 88 M  & 15.4 G & 83.5 \\
\midrule
\multicolumn{4}{l}{\textit{1D SSM}} \\
\midrule
ViM-Ti  \cite{zhuVisionMambaEfficient2024}                 & \phantom{0}7 M   & 1.5 G & 76.1 \\
EfficientViM-M1 \cite{leeEfficientViMHidden2025}  & \phantom{0}7 M   & 0.2 G & 72.9 \\
GroupMamba-T \cite{shakerGroupMambaEfficient2025}   & 23 M  & 4.5 G & 83.3 \\
ViM-S   \cite{zhuVisionMambaEfficient2024}                 & 26 M  & 5.1 G & 80.3 \\
LocalVMamba-T \cite{localmamba}                            & 26 M  & 5.7 G & 82.7 \\
VMamba-T  \cite{liuVMambaVisualState2024}                  & 30 M  & 4.9 G & 82.6 \\
MambaVision-T  \cite{hatamizadehMambaVisionHybridMambaTransformer2024} & 32 M & 4.4 G & 82.3 \\
EfficientVMamba-B \cite{peiEfficientVMambaAtrousSelective2024}         & 33 M & 4.0 G & 81.8 \\
GroupMamba-S \cite{shakerGroupMambaEfficient2025}   & 34 M  & 7.0 G & \hSM{83.9} \\
MambaVision-S  \cite{hatamizadehMambaVisionHybridMambaTransformer2024} & 50 M & 7.5 G & 83.3 \\
VMamba-S  \cite{liuVMambaVisualState2024}                  & 50 M  & 8.7 G & 83.6 \\
GroupMamba-B \cite{shakerGroupMambaEfficient2025}   & 57 M  & 14.0 G & 84.5 \\
VMamba-B  \cite{liuVMambaVisualState2024}                  & 89 M  & 15.4 G & 83.9 \\
MambaVision-B  \cite{hatamizadehMambaVisionHybridMambaTransformer2024} & 98 M & 15.0 G & 84.2 \\
\bottomrule
\toprule
\multicolumn{4}{l}{\textit{Convolutional}} \\
\midrule
ConvNeXt-T   \cite{liuConvNet2020s2022}                    & 29 M  & 4.5 G & 82.1 \\
ConvNeXt-S   \cite{liuConvNet2020s2022}                    & 50 M  & 8.7 G & 83.1 \\
ConvNeXt-B   \cite{liuConvNet2020s2022}                    & 89 M  & 15.4 G & 83.8 \\
\midrule
\multicolumn{4}{l}{\textit{Conv + Attention}} \\
\midrule
CAFormer-S18 \cite{yuMetaFormerBaselinesVision2024}        & 26 M  & 4.1 G & 83.6 \\
CAFormer-M36 \cite{yuMetaFormerBaselinesVision2024}        & 56 M  & 13.2 G & \hSL{85.2} \\
\midrule
\multicolumn{4}{l}{\textit{2D SSM}} \\
\midrule
V2M-T        \cite{wangV2MVisual2Dimensional2024}          & \phantom{0}7 M   & 1.9 G & \hTS{76.2} \\
VSSD-M       \cite{shiVSSDVisionMamba2024arxiv}                 & 14 M  & 2.3 G & \hFS{82.5} \\
VSSD-T       \cite{shiVSSDVisionMamba2024arxiv}                 & 24 M  & 4.5 G & \hTM{83.7} \\
V2M-S        \cite{wangV2MVisual2Dimensional2024}          & 26 M  & 5.9 G & 80.5 \\
S4ND-ConvNeXt-T \cite{nguyenS4NDModelingImages2022}        & 30 M  & -- & 82.2 \\
2DVMamba-T   \cite{zhang2DMambaEfficientState2024}         & 30 M  & 4.9 G & 82.8 \\
V2M-S*       \cite{wangV2MVisual2Dimensional2024}          & 30 M  & 5.4 G & 82.9 \\
VSSD-S       \cite{shiVSSDVisionMamba2024arxiv}                 & 40 M  & 7.4 G & 84.1 \\
2DVMamba-S   \cite{zhang2DMambaEfficientState2024}         & 50 M  & 8.8 G & 83.8 \\
V2M-B*       \cite{wangV2MVisual2Dimensional2024}          & 50 M  & 9.6 G & 83.8 \\
VSSD-B       \cite{shiVSSDVisionMamba2024arxiv}                 & 89 M  & 16.1 G & \hTL{84.7} \\
\midrule
\textbf{M2D-N}                                             & \phantom{0}7 M   & 1.4 G & \hSS{79.6} \\
\textbf{M2D-T}                                             & 27 M  & 4.7 G & \hFM{84.0} \\
\textbf{M2D-S}                                             & 50 M  & 13.7 G & \hFL{85.3} \\
\bottomrule
\end{tabular}%
}%
\end{table}

\paragraph{Block Structure.} We employ a MetaFormer-style \cite{yuMetaFormerActuallyWhat2022} block structure, transforming an input batch $x\in \mathbb{R}^{B \times H \times W \times C}$ to activations $y\in \mathbb{R}^{B \times H \times W \times C}$ as follows:
\begin{equation}
\begin{aligned}
& \hat{x}=\operatorname{Mixer}\left(\operatorname{Norm}\left(x\right)\right)+x, \\
& y =\operatorname{MLP}\left(\operatorname{Norm}\left(\hat{x}\right)\right)+\hat{x},
\end{aligned}
\end{equation}
where $\operatorname{Norm}$ and $\operatorname{Mixer}$ denote our choices of normalisation layer (\ie LayerNorm) and token-mixer (discussed below) respectively, while $\operatorname{MLP}$ denotes a 2-layer MLP with a GELU non-linearity.

\paragraph{Mamba2D Mixer.} We adapt the Mamba token-mixer \cite{guMambaLinearTimeSequence2023}
for visual tasks by removing the symmetric gating branch, which is rendered redundant by
M2D-SSM's own selectivity. In its place, we introduce a local processing path using a depthwise
separable convolution. Similarly to the 1D convolution used in \cite{hatamizadehMambaVisionHybridMambaTransformer2024}, this provides local spatial priors that complement long-range SSM modelling
and improve model convergence.

\paragraph{Overall Architecture.} We implement the Mamba2D mixer described above as a choice of token mixer
within a hierarchical structure adopting the CAFormer
\cite{yuMetaFormerBaselinesVision2024} configurations (see \cref{fig:2DM-model-arch}).
As such, our overall model consists of four stages comprising a strided convolution-based stem/downsample
followed by $N_{1\ldots4}$ mixer blocks within each stage and a post-stage LayerNorm. We ablate the hybrid stage configuration in
\cref{sec:abl-stage-mixer}.
For the task of classification, we employ
Global Average Pooling (GAP)
before feeding the features to an MLP classification head.

\begin{figure}[tb]
\centering
\includegraphics[width=0.93\linewidth]{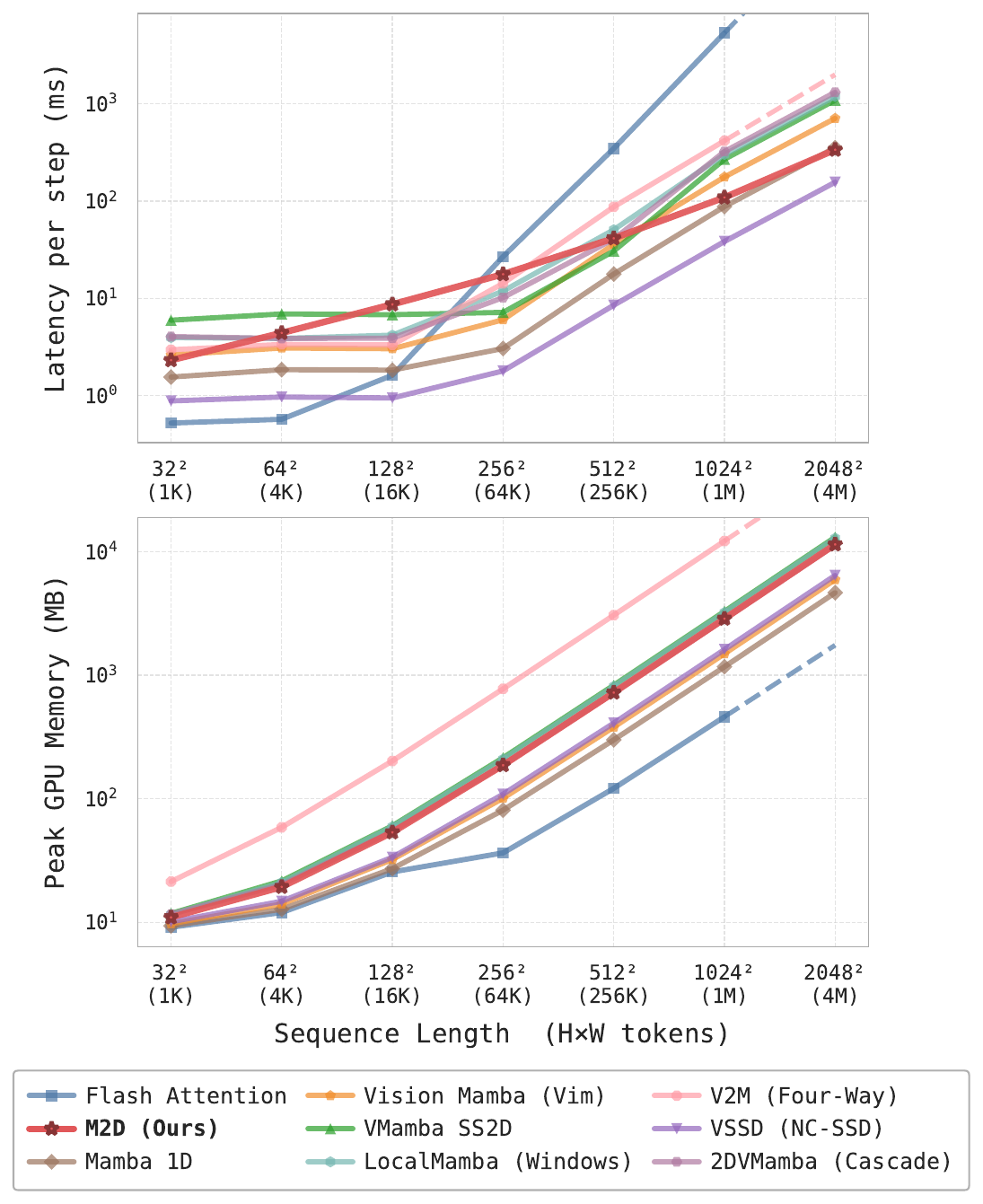}
\caption{Kernel-level scaling of nine token-mixing operators across spatial resolutions ($32^2$--$2048^2$).
\textit{Top:} forward-pass latency (ms, log scale). \textit{Bottom:} peak GPU memory (MB, log scale).
Dashed lines denote log-log extrapolations: Flash Attention~\cite{flashattn} is capped at $1024^2$ due to intractable runtime; V2M runs out of memory at $2048^2$.}
\label{fig:scan-op-scaling}
\vspace{-0.4cm}
\end{figure}

\begin{table}[!b]

\centering
\caption{Peak VRAM and wall time for wavefront-scan recomputation strategies, relative to the naive
CUDA baseline. Sequential$^{\dagger}$ is a kernel-free correctness reference. Full per-strategy detail
is in supplementary \cref{sec:supp-recomp}.}
\label{tab:recomp-compact}
\resizebox{0.9\linewidth}{!}{%
\footnotesize{\setlength{\tabcolsep}{5pt}
\begin{tabular}{l|c|c|c}
\toprule
Strategy & Peak VRAM (\%) & Fwd (\%) & Fwd+Bwd (\%) \\
\midrule
Sequential$^{\dagger}$ & 180 & ${\sim}9.7{\times}10^{3}$ & ${\sim}1.1{\times}10^{6}$ \\
Naive & 100 & 100 & 100 \\
\midrule
Partial recomp. & \textbf{19} & \textbf{46} & \textbf{63} \\
Full recomp. & \textbf{10} & 48 & 65 \\
\bottomrule
\end{tabular}}}
\end{table}

\section{Experiments}
\label{sec:results}

We train the model variants in \cref{tab:model-variants} on
ImageNet-1K~\cite{imagenet} for 300 epochs following standard
practices~\cite{touvronTrainingDataefficientImage2021a,liuSwinTransformerHierarchical2021,liuVMambaVisualState2024}.
Full training details are provided in supplementary \cref{sec:supp-training}.
We additionally evaluate on MS-COCO \cite{COCO} object
detection and ADE20K \cite{ade20k} segmentation.

\begin{table*}[!t]
\centering
\caption{Object detection and instance segmentation results on MS-COCO with Mask R-CNN using 1$\times$ schedule. FLOPs are estimated at input resolution $1280 \times 800$. Within each model size group we highlight the \hFS{1st}\,/\,\hSS{2nd}\,/\,\hTS{3rd} best; highlight width reflects model scale.}
\label{tab:res-coco-1x}
\resizebox{0.75\linewidth}{!}{%
\footnotesize{\setlength{\tabcolsep}{4pt}\renewcommand{\arraystretch}{0.96}
\begin{tabular}{c|ccc|ccc|cc}
\toprule
Backbone & $\mathrm{AP}^{\mathrm{b}}$ & $\mathrm{AP}_{50}^{\mathrm{b}}$ & $\mathrm{AP}_{75}^{\mathrm{b}}$ & $\mathrm{AP}^{\mathrm{m}}$ & $\mathrm{AP}_{50}^{\mathrm{m}}$ & $\mathrm{AP}_{75}^{\mathrm{m}}$ & \#Param. & FLOPs \\
\midrule
PVT-T \cite{wangPyramidVisionTransformer2021} & 36.7 & 59.2 & 39.3 & 35.1 & 56.7 & 37.3 & 33 M & 208 G \\
EffVMamba-S \cite{peiEfficientVMambaAtrousSelective2024} & 39.3 & 61.8 & 42.8 & 36.7 & 58.9 & 39.2 & 31 M & 197 G \\
\textbf{M2D-N} & \hTS{42.8} & \hTS{65.3} & \hTS{46.7} & \hTS{39.6} & \hTS{62.5} & \hTS{42.5} & 26 M & 175 G \\
MSVMamba-M \cite{MSVMamba} & \hSS{43.8} & \hSS{65.8} & \hSS{47.7} & \hSS{39.9} & \hSS{62.9} & \hSS{42.9} & 32 M & 201 G \\
VSSD-M \cite{shiVSSDVisionMamba2024arxiv} & \hFS{45.4} & \hFS{67.5} & \hFS{49.8} & \hFS{41.3} & \hFS{64.5} & \hFS{44.6} & 33 M & 220 G \\
\midrule
ResNet-50 \cite{resnet} & 38.2 & 58.8 & 41.4 & 34.7 & 55.7 & 37.2 & 44 M & 260 G \\
Swin-T \cite{liuSwinTransformerHierarchical2021} & 42.7 & 65.2 & 46.8 & 39.3 & 62.2 & 42.2 & 48 M & 267 G \\
ConvNeXt-T \cite{liuConvNet2020s2022} & 44.2 & 66.6 & 48.3 & 40.1 & 63.3 & 42.8 & 48 M & 262 G \\
VSSD-T \cite{shiVSSDVisionMamba2024arxiv} & 46.9 & \hTM{69.4} & 51.4 & 42.6 & \hTM{66.4} & \hTM{45.9} & 44 M & 265 G \\
VMamba-T \cite{liuVMambaVisualState2024} & \hTM{47.3} & 69.3 & 52.0 & \hTM{42.7} & \hTM{66.4} & \hTM{45.9} & 50 M & 271 G \\
V2M-S \cite{wangV2MVisual2Dimensional2024} & \hSM{47.6} & \hTM{69.4} & \hSM{52.2} & \hSM{42.9} & \hSM{66.5} & \hSM{46.3} & 50 M & - \\
GroupMamba-T \cite{shakerGroupMambaEfficient2025} & \hSM{47.6} & \hSM{69.8} & \hTM{52.1} & \hSM{42.9} & \hSM{66.5} & \hSM{46.3} & 40 M & 279 G \\
\textbf{M2D-T} & \hFM{48.5} & \hFM{71.0} & \hFM{53.1} & \hFM{43.8} & \hFM{67.8} & \hFM{47.1} & 44 M & 288 G \\
\midrule
ResNet-101 \cite{resnet} & 40.4 & 61.1 & 44.2 & 36.4 & 58.1 & 38.8 & 63 M & 336 G \\
Swin-S \cite{liuSwinTransformerHierarchical2021} & 44.8 & 66.6 & 48.9 & 40.9 & 63.2 & 44.2 & 69 M & 354 G \\
ConvNeXt-S \cite{liuConvNet2020s2022} & 45.4 & 67.9 & 50.0 & 41.8 & 65.0 & 45.1 & 70 M & 348 G \\
VSSD-S \cite{shiVSSDVisionMamba2024arxiv} & 48.4 & \hTL{70.1} & 53.1 & 43.5 & 67.2 & \hSL{47.1} & 59 M & 325 G \\
VMamba-S \cite{liuVMambaVisualState2024} & \hTL{48.7} & 70.0 & \hTL{53.4} & \hTL{43.7} & \hTL{67.3} & \hTL{47.0} & 70 M & 349 G \\
V2M-B \cite{wangV2MVisual2Dimensional2024} & \hSL{48.9} & \hSL{70.2} & \hSL{53.6} & \hSL{43.8} & \hSL{67.5} & \hSL{47.1} & 70 M & - \\
\textbf{M2D-S} & \hFL{50.4} & \hFL{72.3} & \hFL{55.3} & \hFL{45.1} & \hFL{69.4} & \hFL{48.6} & 67 M & 549 G \\
\bottomrule
\end{tabular}}}
\end{table*}

\subsection{Image Classification on ImageNet-1K}

\Cref{tab:class-res} presents ImageNet-1K classification results. M2D reaches state-of-the-art
accuracy at every scale: M2D-T attains 84.0\%, surpassing the previous best 2D-SSM baseline and all
comparable 1D-SSM and attention-based models, including several larger architectures, and M2D-S
attains 85.3\%, ahead of the strongest SSM baselines by clear margins. \Cref{sec:abl-m1d-main}
examines how much of this performance owes to the 2D scan itself rather than the hybrid design
around it.

\subsection{Computational Efficiency and Scalability}

\label{sec:scan-op-comparison}

As shown in \cref{fig:scan-op-scaling}, M2D scales sub-linearly in latency due to the wavefront's
diagonal-depth growth, and at high resolution is substantially faster than Flash
Attention~\cite{flashattn}, VMamba (SS2D) and V2M, while all SSM-based operators maintain linear memory
scaling. The fusion and recomputation strategies of \cref{sec:cuda-wf-impl} exploit the memory-bound
nature of the scan (\cref{tab:recomp-compact}): partial recomputation reduces peak VRAM by
${\sim}81\%$ while also reducing wall time, and full recomputation reaches ${\sim}10\%$ of the naive
footprint at a small latency cost. The
full operator comparison and per-strategy breakdowns are in supplementary \cref{sec:supp-scan-op,sec:supp-recomp}.

\subsection{Object Detection and Instance Segmentation}
For object detection and instance segmentation, we fine-tune
Mask R-CNN \cite{Detectron2018} on MS-COCO \cite{COCO} using
MMDetection \cite{mmdet} under 1${\times}$ and 3${\times}$
schedules. Full training details are in supplementary \cref{sec:supp-training}.

As shown in \cref{tab:res-coco-1x,tab:res-coco-3x}, M2D-T and M2D-S set a new state-of-the-art
among 2D-SSM backbones at both schedules.
M2D-N
remains competitive at substantially lower FLOPs than alternatives in its group. M2D-S's higher
FLOPs, by contrast, stem from allocating more blocks to the high-resolution stage, where Mask R-CNN
concentrates its compute, a trade-off that inverts on ADE20K (supplementary \cref{sec:supp-flops}). MambaVision, the
closest hybrid baseline, evaluates detection with the stronger Cascade Mask R-CNN head and is
therefore incomparable with these results. Under the plain Mask R-CNN 1$\times$ protocol, its own
ablation reports 46.4 AP$^{b}$, 2.1 points below M2D-T.

\begin{table*}[!t]
\centering
\caption{Object detection \& instance segmentation results on MS-COCO with Mask R-CNN-3$\times$ multi-scale schedule. FLOPs are estimated at input resolution $1280 \times 800$. $^\dagger$CAFormer uses sliding-window attention for detection, as full attention exceeds memory \cite{yuMetaFormerBaselinesVision2024}. Within each model size group we highlight the \hFS{1st}\,/\,\hSS{2nd}\,/\,\hTS{3rd} best; highlight width reflects model scale.}
\label{tab:res-coco-3x}
\resizebox{0.75\linewidth}{!}{%
\footnotesize{\setlength{\tabcolsep}{4pt}\renewcommand{\arraystretch}{0.96}
\begin{tabular}{c|ccc|ccc|cc}
\toprule
Backbone & $\mathrm{AP}^{\mathrm{b}}$ & $\mathrm{AP}_{50}^{\mathrm{b}}$ & $\mathrm{AP}_{75}^{\mathrm{b}}$ & $\mathrm{AP}^{\mathrm{m}}$ & $\mathrm{AP}_{50}^{\mathrm{m}}$ & $\mathrm{AP}_{75}^{\mathrm{m}}$ & \#Param. & FLOPs \\
\midrule
PVT-T \cite{wangPyramidVisionTransformer2021} & 39.8 & 62.2 & 43.0 & 37.4 & 59.3 & 39.9 & 33 M & 208 G \\
LightViT-T \cite{lightvit} & 41.5 & 64.4 & 45.1 & 38.4 & 61.2 & 40.8 & 28 M & 187 G \\
EffVMamba-S \cite{peiEfficientVMambaAtrousSelective2024} & 41.6 & 63.9 & 45.6 & 38.2 & 60.8 & 40.7 & 31 M & 197 G \\
MSVMamba-M \cite{MSVMamba} & \hSS{46.3} & \hSS{68.1} & \hSS{50.8} & \hSS{41.8} & \hSS{65.1} & \hSS{44.9} & 32 M & 201 G \\
\textbf{M2D-N} & \hTS{47.2} & \hTS{69.2} & \hTS{51.7} & \hTS{42.4} & \hTS{65.9} & \hTS{45.6} & 26 M & 175 G \\
VSSD-M \cite{shiVSSDVisionMamba2024arxiv} & \hFS{47.7} & \hFS{69.7} & \hFS{52.1} & \hFS{42.8} & \hFS{66.5} & \hFS{46.0} & 33 M & 220 G \\
\midrule
ResNet-50 \cite{resnet} & 41.0 & 61.7 & 44.9 & 37.1 & 58.4 & 40.1 & 44 M & 260 G \\
Swin-T \cite{liuSwinTransformerHierarchical2021} & 46.0 & 68.1 & 50.3 & 41.6 & 65.1 & 44.9 & 48 M & 267 G \\
ConvNeXt-T \cite{liuConvNet2020s2022} & 46.2 & 67.9 & 50.8 & 41.7 & 65.0 & 44.9 & 48 M & 262 G \\
CAFormer-S18$^\dagger$ \cite{yuMetaFormerBaselinesVision2024} & 48.6 & \hTM{70.5} & \hTM{53.4} & \hTM{43.7} & \hTM{67.5} & \hSM{47.4} & -- & 254 G \\
VSSD-T \cite{shiVSSDVisionMamba2024arxiv} & \hTM{48.8} & 70.4 & \hTM{53.4} & 43.6 & \hSM{67.6} & 46.9 & 44 M & 265 G \\
VMamba-T \cite{liuVMambaVisualState2024} & \hTM{48.8} & 70.4 & \hSM{53.5} & \hTM{43.7} & 67.4 & 47.0 & 50 M & 271 G \\
V2M-S* \cite{wangV2MVisual2Dimensional2024} & \hSM{49.0} & \hSM{70.6} & \hSM{53.5} & \hSM{43.8} & \hTM{67.5} & \hTM{47.2} & 50 M & - \\
\textbf{M2D-T} & \hFM{50.5} & \hFM{72.1} & \hFM{55.4} & \hFM{44.7} & \hFM{68.9} & \hFM{48.3} & 44 M & 288 G \\
\midrule
ResNet-101 \cite{resnet} & 42.8 & 63.2 & 47.1 & 38.5 & 60.1 & 41.4 & 63 M & 336 G \\
Swin-S \cite{liuSwinTransformerHierarchical2021} & 48.2 & \hTL{69.8} & 52.8 & 43.2 & 67.0 & 46.1 & 69 M & 354 G \\
ConvNeXt-S \cite{liuConvNet2020s2022} & 47.9 & 69.0 & 52.7 & 42.9 & 66.9 & 46.2 & 70 M & 348 G \\
VMamba-S \cite{liuVMambaVisualState2024} & \hTL{49.9} & \hSL{70.9} & \hTL{54.7} & \hTL{44.2} & \hTL{68.2} & \hTL{47.7} & 70 M & 349 G \\
V2M-B* \cite{wangV2MVisual2Dimensional2024} & \hSL{50.0} & \hSL{70.9} & \hSL{54.8} & \hSL{44.3} & \hSL{68.4} & \hSL{47.8} & 70 M & - \\
\textbf{M2D-S} & \hFL{52.2} & \hFL{73.1} & \hFL{57.4} & \hFL{46.2} & \hFL{70.6} & \hFL{50.4} & 67 M & 549 G \\
\bottomrule
\end{tabular}}}
\vspace{-0.2cm}
\end{table*}

\subsection{Semantic Segmentation}

For semantic segmentation on ADE20K \cite{ade20k}, we employ
UperNet \cite{UperNet} with MMSegmentation \cite{mmseg}, evaluating under
both single-scale (SS) and multi-scale (MS) settings.
Full training details are in supplementary \cref{sec:supp-training}.
As shown in \cref{tab:res-ade-20k}, M2D-T and M2D-S set a new state-of-the-art among SSM-based
approaches, reaching 48.9 and 51.7 SS mIoU at 485G and 768G FLOPs respectively, with M2D-N
competitive at 355G. Both larger variants retain their leads under multi-scale evaluation.

\vspace{-0.2cm}
\section{Ablations}
\label{sec:ablations}
\vspace{-0.1cm}

To separate the contributions of our individual design choices, we ablate M2D-N following the training setup of \cref{sec:results}.
Following standard practice, high-level architectural decisions (stage-wise mixer selection) are evaluated on
ImageNet-1K classification accuracy and throughput. Design choices with broader impact (\eg component contributions and 
scan patterns) are additionally validated on downstream dense prediction benchmarks.
All throughput measurements use bfloat16 inference on an NVIDIA RTX 5090.

\subsection{Stage-wise Token Mixer Selection}
\label{sec:abl-stage-mixer}

\begin{figure}[!t]
\centering
\includegraphics[width=0.93\linewidth]{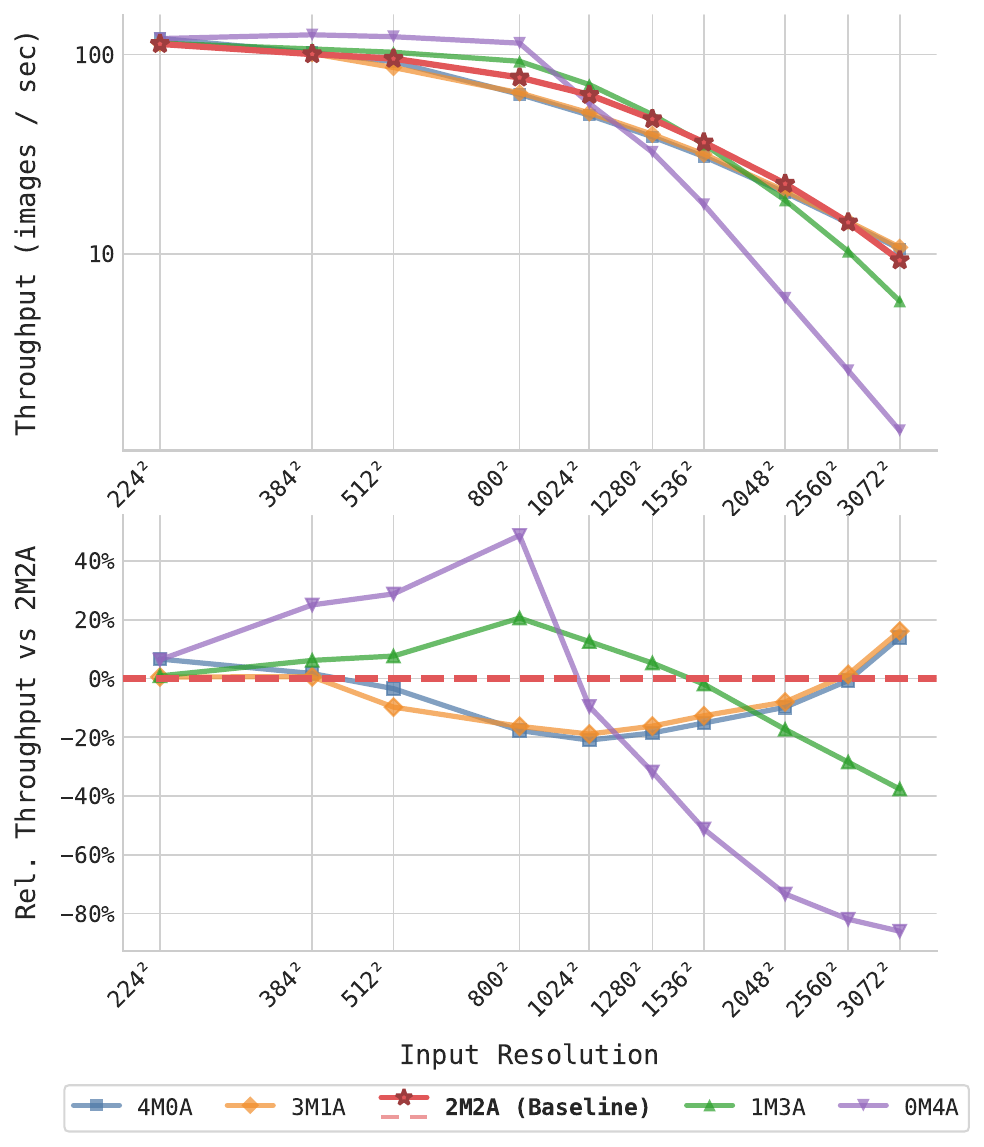}
\caption{Throughput vs input resolution for different stage-wise mixer configurations. \textbf{Top:} Absolute throughput on log-log scale. \textbf{Bottom:} Relative performance vs 2M2A baseline.
}
\label{fig:scaling}
\end{figure}

\Cref{tab:abl-stage-mixer} and \cref{fig:scaling} report ImageNet-1K accuracy and throughput for M2D-N
with varying proportions of Mamba2D (M) and attention (A) stages across the four network stages;
configurations are denoted $xM\,yA$.

Accuracy saturates beyond 2M2A (79.6\%), with 1M3A adding only 0.1\% at significant
throughput cost. The crossover at input size $1536^2$ (\cref{fig:scaling}, right) shows 2M2A surpassing all
higher-attention configurations. At $2048^2$ inputs, it is 21\% faster than 1M3A and $3.8\times$ faster than
0M4A. This motivates the 2M2A configuration used throughout: restricting attention to later stages, where
spatial dimensions are small, captures its modelling benefits without incurring quadratic cost.

\subsection{Mamba2D Mixer Components}

\Cref{tab:abl-mixer-components} evaluates the contribution of each parallel path in the Mamba2D mixer:
the native 2D SSM path and the local depthwise separable convolution path.

Image-level classification is known to be largely insensitive to the token mixer: MambaOut
\cite{mambaout} is competitive with no SSM at all, and RandFormer
\cite{yuMetaFormerBaselinesVision2024} reaches 76.6--81.4\% top-1 with random mixing. Both paths
accordingly contribute minimally here, while being
complementary on dense prediction. The local path captures texture and boundary
information (the primary signal for per-pixel segmentation), explaining the larger 1.3\,mIoU drop
when it is removed. Detection requires integrating spatial context across larger regions to localise
instances, where the long-range M2D scan is better suited, causing a 1.7\,AP$^b$ drop
when removed, compared to only 0.9\,AP$^b$ when the local path is removed. Removing the M2D path entirely
marginally outperforms retaining the projections without the scan,
confirming that the SSM projections are only useful alongside the scan. 

\begin{table}[!t]
\centering
\caption{Semantic segmentation results on ADE20K with UperNet. FLOPs are estimated at crop size $512 \times 2048$.
}
\label{tab:res-ade-20k}
\resizebox{\linewidth}{!}{%
\small{\setlength{\tabcolsep}{8pt}
\begin{tabular}{c|cc|cc}
\toprule
Backbone & \#Param. & FLOPs & mIoU (SS) & mIoU (MS) \\
\midrule
Vim-Ti \cite{zhuVisionMambaEfficient2024}              & 13 M  & --     & 41.0 & --   \\
V2M-T \cite{wangV2MVisual2Dimensional2024}              & 13 M  & --     & 41.4 & 42.0 \\
EfficientVMamba-S \cite{peiEfficientVMambaAtrousSelective2024}  & 29 M  & 505 G  & \hTS{41.5} & \hTS{42.1} \\
\textbf{M2D-N}      & 34 M  & 355 G  & \hSS{43.1} & \hSS{43.1} \\
LocalVim-T \cite{localmamba}         & 36 M  & 260 G  & \hFS{43.4} & \hFS{44.4} \\
\midrule
ResNet-50 \cite{resnet} & 66 M  & 953 G  & 42.1 & 42.8 \\
Swin-T \cite{liuSwinTransformerHierarchical2021}             & 60 M  & 945 G  & 44.4 & 45.8 \\
Vim-S \cite{zhuVisionMambaEfficient2024}              & 46 M  & --     & 44.9 & --   \\
V2M-S \cite{wangV2MVisual2Dimensional2024}              & 46 M  & 482 G  & 45.1 & 46.1 \\
MSVMamba-M \cite{MSVMamba}  & 42 M  & 875 G  & 45.1 & 45.4 \\
VSSD-M \cite{shiVSSDVisionMamba2024arxiv}             & 42 M  & 893 G  & 45.6 & 46.0 \\
EffVMamba-B \cite{peiEfficientVMambaAtrousSelective2024}        & 65 M  & 930 G  & 46.5 & 47.3 \\
MSVMamba-T \cite{MSVMamba}  & 65 M  & 942 G  & 47.6 & 48.5 \\
MambaVision-T \cite{hatamizadehMambaVisionHybridMambaTransformer2024} & 55 M & 945 G & 46.0 & -- \\
VSSD-T \cite{shiVSSDVisionMamba2024arxiv}             & 53 M  & 941 G  & 47.9 & 48.7 \\
VMamba-T \cite{liuVMambaVisualState2024}           & 62 M  & 948 G  & 47.9 & 48.8 \\
V2M-S \cite{wangV2MVisual2Dimensional2024}             & 62 M  & 482 G  & \hTM{48.2} & \hTM{49.0} \\
GroupMamba-T \cite{shakerGroupMambaEfficient2025}      & --    & --     & \hSM{48.6} & \hSM{49.2} \\
CAFormer-S18 \cite{yuMetaFormerBaselinesVision2024}    & 54 M  & 1024 G & --   & 48.9 \\
\textbf{M2D-T}      & 53 M  & 485 G  & \hFM{{48.9}} & \hFM{{49.3}} \\
\midrule
ResNet-101 \cite{resnet} & 85 M  & 1030 G & 44.9 & 45.9 \\
Swin-S \cite{liuSwinTransformerHierarchical2021}             & 81 M  & 1039 G & 47.6 & 49.5 \\
NAT-S \cite{NAT}    & 82 M  & 1010 G & 48.0 & 49.5 \\
2DMamba \cite{zhang2DMambaEfficientState2024} & 92 M  & 950 G  & 48.6 & 49.3 \\
ConvNeXt-S \cite{liuConvNet2020s2022}         & 82 M  & 1028 G & 48.7 & 49.6 \\
MambaVision-S \cite{hatamizadehMambaVisionHybridMambaTransformer2024} & 84 M & 1135 G & 48.2 & -- \\
MambaVision-B \cite{hatamizadehMambaVisionHybridMambaTransformer2024} & 126 M & 1342 G & 49.1 & -- \\
VMamba-S \cite{liuVMambaVisualState2024}           & 82 M  & 1028 G & \hTL{50.6} & 51.2 \\
V2M-B \cite{wangV2MVisual2Dimensional2024}             & 82 M  & 655 G  & \hSL{50.8} & \hTL{51.3} \\
CAFormer-M36 \cite{yuMetaFormerBaselinesVision2024}    & 84 M  & 1346 G & --   & \hSL{51.7} \\
\textbf{M2D-S}      & 77 M  & 768 G  & \hFL{{51.7}} & \hFL{{51.8}} \\
\bottomrule
\end{tabular}}}
\vspace{-0.6cm}
\end{table}

\subsection{Scan Direction Strategies}
\label{sec:abl-scan-dir}

We evaluate scanning strategies that vary the number of directions and whether directions are
alternated between blocks (``alternating'') or duplicated within each block 
(``per block'').
\Cref{tab:abl-scan-dir} reports accuracy and throughput.

\begin{table}
\centering
\caption{Ablation of stage-wise token mixer selection on ImageNet-1K (M2D-N). 
}
\label{tab:abl-stage-mixer}

\scriptsize{\setlength{\tabcolsep}{4pt}
\begin{tabular}{l|ccccc}
\toprule
 & 4M 0A & 3M 1A & \textbf{2M 2A} & 1M 3A & 0M 4A \\
\midrule
Acc. (\%) & \cii{78.6} & \cii{78.9} & \hS{\cii{79.6}} & \hF{\cii{79.7}} & \hT{\cii{79.4}} \\
\#Param. (M) & \cii{8.4} & \cii{7.9} & \cii{7.2} & \cii{7.1} & \cii{7.1} \\
\midrule
\multicolumn{6}{c}{Throughput (images/sec)} \\
\midrule
224$^2$  & \hF{\ciii{120.7}} & \ciii{113.8} & \ciii{113.2} & \hT{\ciii{114.2}} & \hS{\ciii{120.4}} \\
512$^2$  & \cii{92.2} & \cii{86.2} & \hT{\cii{95.5}} & \hS{\ciii{102.8}} & \hF{\ciii{123.0}} \\
1024$^2$ & \cii{49.7} & \cii{51.0} & \hS{\cii{62.9}} & \hF{\cii{70.8}} & \hT{\cii{56.9}} \\
1536$^2$ & \cii{30.8} & \hT{\cii{31.7}} & \hF{\cii{36.3}} & \hS{\cii{35.6}} & \cii{17.7} \\
2048$^2$ & \hT{\cii{20.3}} & \hS{\cii{20.7}} & \hF{\cii{22.5}} & \cii{18.6} & \ci{6.0} \\
2560$^2$ & \hT{\cii{14.3}} & \hF{\cii{14.6}} & \hS{\cii{14.4}} & \cii{10.3} & \ci{2.6} \\
3072$^2$ & \hS{\cii{10.6}} & \hF{\cii{10.8}} & \hT{\ci{9.3}} & \ci{5.8} & \ci{1.3} \\
\bottomrule
\end{tabular}}
\end{table}

\begin{table}[!t]
\centering
\caption{Ablation of Mamba2D mixer components. 
}
\label{tab:abl-mixer-components}
\resizebox{1.0\linewidth}{!}{%
\footnotesize{\setlength{\tabcolsep}{4pt}
\begin{tabular}{l|c|c|c|c}
\toprule
Configuration & Top-1 (\%) & AP$^b$ & AP$^m$ & mIoU \\
\midrule
Baseline (full) & \hF{79.6} & \hF{42.8} & \hF{39.6} & \hF{43.1} \\
w/o local path & \hS{79.4} & \hS{41.9} & \hS{39.0} & 41.8 \\
w/o M2D scan & \hF{79.6} & 41.1 & \hT{38.4} & \hS{42.8} \\
w/o M2D path (conv only) & \hS{79.4} & \hT{41.3} & \hT{38.4} & \hT{42.6} \\
\bottomrule
\end{tabular}}}
\vspace{-0.3cm}
\end{table}

\begin{table}[t]
\centering
\caption{Ablation of scan direction strategies.
Throughput is normalised to the baseline. 
}
\label{tab:abl-scan-dir}
\resizebox{0.97\linewidth}{!}{%
{\setlength{\tabcolsep}{5pt}
\begin{tabular}{l|c|c|c|c|c}
\toprule
Scan Strategy & Throughput & Top-1 (\%) & AP$^b$ & AP$^m$ & mIoU \\
\midrule
TL$\rightarrow$BR only                              & \hF{1.01$\times$} & \hS{79.8} & 41.8 & 38.9 & \hT{42.9} \\
2-dir alternating (baseline)                        & \hS{1.00$\times$} & 79.6 & \hS{42.8} & \hS{39.6} & \hS{43.1} \\
2-dir per block                                     & \hT{\kern0.62pt0.65$\times$\kern0.62pt} & \hT{79.7} & \hF{43.0} & \hF{39.8} & 42.7 \\
4-dir alternating                                   & \hS{1.00$\times$} & \hT{79.7} & \hT{42.7} & \hT{39.5} & \hT{42.9} \\
4-dir per block                                     & 0.38$\times$ & \hF{80.0} & \hS{42.8} & \hS{39.6} & \hF{43.9} \\
\bottomrule
\end{tabular}}}
\vspace{-0.1cm}
\end{table}

Single-direction scanning (TL$\rightarrow$BR) stays competitive on classification (79.8\%) with a slight detection drop (41.8 vs. 42.8\,AP$^b$). Scanning from different corners varies this by only $\pm0.3$ AP$^b$. 2-dir alternating recovers this at no throughput cost. 4-dir alternating adds
nothing further, as two complementary directions already capture the bidirectional context that dense
prediction requires. Per-block strategies reach marginally higher peak scores
(43.0\,AP$^b$, 43.9\,mIoU), though the throughput penalties ($0.65\times$ and $0.38\times$) are not justified by the modest gains.
We therefore adopt 2-dir alternating as our default.

\subsection{Scan Dimensionality}
\label{sec:abl-m1d-main}

To isolate the contribution of the native 2D scan from the hybrid design, we train a matched
baseline, M1D, replacing the stage-1/2 scan operator with a standard 1D selective scan over the
row-major flattening of the feature map.
Full construction details are in supplementary \cref{sec:abl-m1d}.

\begin{table}[!t]
\centering
\caption{Ablation of scanning strategy using a matched M1D baseline on Imagenet-1K, and MS-COCO ($1\times$ schedule); $\Delta=$M2D$-$M1D. \textbf{Bold} values indicate M2D outperforms M1D.}
\label{tab:abl-m1d}
\resizebox{\linewidth}{!}{%
\scriptsize{\setlength{\tabcolsep}{3pt}
\begin{tabular}{l|r|rrrr|rrrr}
\toprule
\multirow{2}{*}{Scale} & IN-1K &\multicolumn{4}{c|}{\rule{0pt}{2.15ex}bbox} & \multicolumn{4}{c}{\rule{0pt}{2.15ex}segm} \\
\cline{2-10}
 & top1\% & \rule{0pt}{2.15ex}mAP & AP$_S$ & AP$_M$ & AP$_L$ & mAP & AP$_S$ & AP$_M$ & AP$_L$ \\
\midrule
N-scale & $-0.1$ & $\mathbf{+0.1}$ & $\mathbf{+0.5}$ & $\mathbf{+0.1}$ & $\mathbf{+0.4}$ & $0.0$ & $\mathbf{+0.8}$ & $0.0$ & $-0.2$ \\
T-scale & $0.0$ & {$ \mathbf{+0.3}$} & $\mathbf{+1.2}$ & $-0.2$ & $\mathbf{+0.8}$ & $\mathbf{+0.3}$ & $\mathbf{+1.0}$ & $-0.2$ & $\mathbf{+0.3}$ \\
\bottomrule
\end{tabular}}}
\vspace{-0.3cm}
\end{table}

On ImageNet pretraining the two models tie,
consistent with the mixer-insensitivity of classification (\cref{tab:abl-mixer-components}). On dense
prediction M2D pulls ahead (\cref{tab:abl-m1d}), with gains concentrated on small objects (AP$_S$),
where fine spatial adjacency matters most, roughly doubling from N- to T-scale. Replacing the scan instead with VMamba's
cross-scan~\cite{liuVMambaVisualState2024}  yields similar N-scale accuracy, but from
four separately parametrised 1D passes rather than one joint 2D scan, with markedly worse
resolution scaling (\cref{fig:scan-op-scaling}, supplementary \cref{sec:abl-m1d}).
The scan formulation and the hybrid design are therefore complementary contributions.

\section{Conclusion}

In this paper, we introduced Mamba2D, a natively multi-dimensional state-space model for vision that derives a true
joint 2D selective SSM from first principles rather than composing or factoring 1D scans. To make this
tractable, we implement and release a custom CUDA wavefront-scan kernel that exploits diagonal independence for
parallelism, with selective activation recomputation achieving linear memory scaling at high
resolution. Our hybrid architecture, pairing native 2D SSMs in high-resolution stages with
vanilla attention in low-resolution stages, achieves state-of-the-art results among SSM-based models
on ImageNet-1K classification, MS-COCO detection and ADE20K segmentation across model sizes.

A current limitation is that the wavefront's sequential diagonal ordering carries a higher
constant-factor latency than 1D operators at low resolutions, an overhead amortised at the high
resolutions where M2D is most naturally deployed and one that hardware-aware diagonal scheduling
should further reduce. More broadly, the formulation extends beyond 2D, towards natively
volumetric SSMs for video, medical imaging and event-based sensing \cite{emamba}.

\section*{Acknowledgements}

This work was supported by Surrey Satellite Technology Ltd (SSTL) and UK Research and Innovation (UKRI).

{
    \small
    \bibliographystyle{ieeenat_fullname}
    \bibliography{main}
}

\clearpage
\appendix
\renewcommand{\theHsection}{appendix.\arabic{section}}
\twocolumn[{\centering\Large\textbf{Supplementary Material}\par\vspace{1.2em}}]

\begin{table*}[!t]
\centering
\small
\caption{Training and implementation details for ImageNet-1K classification,
MS-COCO object detection, and ADE20K semantic segmentation. Unspecified
hyperparameters follow the respective framework defaults.}
\label{tab:supp-training}
\setlength{\tabcolsep}{5pt}
\resizebox{0.8\textwidth}{!}{%
\begin{tabular}{l|c|c|c}
\toprule
\textbf{Hyperparameter} & \textbf{ImageNet-1K} & \textbf{COCO (det.)} & \textbf{ADE20K (seg.)} \\
\midrule
Framework         & ---                   & MMDetection \cite{mmdet}   & MMSegmentation \cite{mmseg} \\
Head              & GAP + MLP             & Mask R-CNN \cite{Detectron2018} + FPN \cite{FPN} & UperNet \cite{UperNet} \\
Optimiser         & AdamW                 & AdamW                 & AdamW \\
Learning rate     & $4{\times}10^{-3}$    & $1{\times}10^{-4}$    & $6{\times}10^{-5}$ \\
LR schedule       & Cosine annealing      & \makecell[c]{MultiStep (${\times}0.1$ at ep.\,8,11\\/ ep.\,27,33)} & Poly (power 1.0) \\
Warmup            & 5\% linear            & Linear (1k iters)     & Linear (1.5k iters) \\
Weight decay      & 0.05                  & 0.05                  & 0.05 \\
Batch size        & 4096                  & 16                    & 16 \\
Duration          & 300 epochs            & 1${\times}$ / 3${\times}$ schedule & 160k iterations \\
Input resolution  & $224{\times}224$      & $1333{\times}800$     & $512{\times}512$ crop \\
Precision         & bfloat16              & fp32                  & fp32 \\
EMA decay         & 0.999                 & ---                   & --- \\
FPN normalisation & ---                   & Group Norm \cite{groupnorm} & --- \\
Augmentation      & \makecell[c]{RandAug~\cite{randaugment}, flip, crop,\\jitter, erasing~\cite{randerasing},\\Mixup/CutMix~\cite{mixup,cutmix}, LS} & \makecell[c]{Flip (1${\times}$)\\Flip + LSJ~\cite{carion2020detr} (3${\times}$)} & \makecell[c]{Flip, resized crop\\($[0.5,2.0]$), photometric\\distortion} \\
\bottomrule
\end{tabular}
}%
\end{table*}

\section{Training and Implementation Details}
\label{sec:supp-training}

\Cref{tab:supp-training} summarises the training hyperparameters and
implementation details for all three benchmarks. All unspecified
hyperparameters follow the default settings of the respective framework.
Each reported result is a single training run, as is standard practice for ImageNet-scale
backbones (A 300-epoch pretrain of M2D-T alone takes approximately 150 GPU-hours).
Accordingly, we do not draw conclusions from differences of a few tenths of a point in overall mAP,
which we treat as within ordinary run-to-run variation. 

\section{Boundedness and the 1/2 Combination Factor}
\label{sec:supp-alpha}

The M2D recurrence (\cref{sec:method}) estimates each hidden state from two predecessors, one
per spatial axis, and combines the two axis-wise contributions with a fixed factor of $1/2$. Writing
that coupling factor as a general scalar $\alpha$, the recurrence is
\begin{equation}
\begin{split}
h(t,z) ={}& \alpha\left(\overline{\mathbf{A}}_t\, h(t-\Delta_t,z) + \overline{\mathbf{A}}_z\, h(t,z-\Delta_z)\right)\\
&+ \alpha\left(\overline{\mathbf{B}}_t + \overline{\mathbf{B}}_z\right)x(t,z),
\end{split}
\label{eqn:supp-alpha-rec}
\end{equation}
recovering the M2D formulation at $\alpha = 1/2$. 2D-SSM \cite{baron2DimensionalStateSpace2024} uses the same factor in its two-axis recurrence. This section shows that $1/2$ is the largest value that prevents unbounded growth, then discusses learned versions of $\alpha$
and finally demonstrates the effect of varying $\alpha$ empirically.

\paragraph{The growth bound.}
Each hidden state is formed from two predecessors, one per spatial axis. Writing
$\rho = \max(\|\overline{\mathbf{A}}_t\|, \|\overline{\mathbf{A}}_z\|)$ for the larger of the two
transition norms, the state propagated onto an anti-diagonal is therefore amplified by at most
$2\alpha\rho$ relative to the preceding anti-diagonal (the two contributions are each scaled by at most
$\alpha\rho$). Under the standard Mamba parametrisation $\mathbf{A}$ is diagonal with negative
entries and $\Delta > 0$, so $\overline{\mathbf{A}} = \exp(\Delta\mathbf{A})$ has entries in
$(0,1]$ and $\rho \leq 1$ for every input-dependent parameter value. At $\alpha = 1/2$ the
worst-case per-step gain is thus at most $1$: earlier states are never amplified, and the
recurrence cannot grow exponentially across the grid. Any larger $\alpha$ can lead to exploding values, and requires the learned $\mathbf{A}$ matrix to take responsibility for keeping the state bounded, by satisfying $\rho < 1/(2\alpha)$. 
Values for $\alpha$ smaller than $1/2$ also do not lead to amplified values, but they forcibly scale
the state down at every step, shortening the effective memory unnecessarily. With
two predecessors, $\alpha = 1/2$ is thus the largest value that prevents unbounded growth, and
hence the one that damps the hidden state least.

A learned coefficient sounds initially appealing. However, per the above analysis, our system could not learn any globally applied value of $\alpha$ that would improve the effective memory beyond that of $\alpha=0.5$ without re-introducing the possibility of instability. We might also consider an input-dependent alpha, modulating the combination factor in different regions of the image. However, such an extension actually adds no additional expressivity beyond the global $\alpha$. The recurrence depends on $\alpha$ and the discretised parameters only through the products
$\alpha\overline{\mathbf{A}}_{t,z}$ and $\alpha\overline{\mathbf{B}}_{t,z}$, so $\alpha$ is not an
independent degree of freedom, and per-element control of the same quantities is already encoded
into the input-dependent $\overline{\mathbf{A}}_{t,z}$ and $\overline{\mathbf{B}}_{t,z}$. Any learned adaptation of $\alpha$ would either re-introduce the risk of unbounded growth, or duplicate control the model already has.

\begin{figure}[!htb]
\centering
\includegraphics[width=\linewidth]{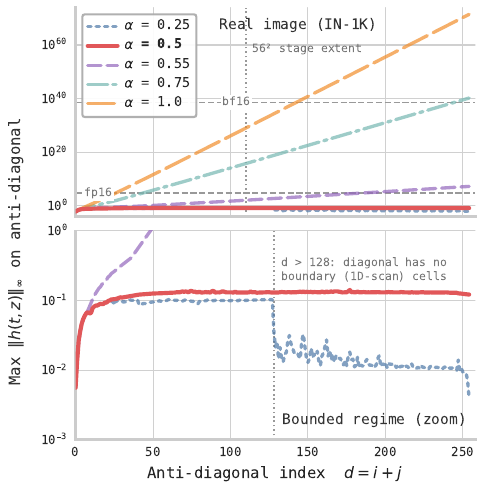}
\caption{Hidden-state norm growth at initialisation for coupling factors
$\alpha \in \{0.25, 0.5, 0.55, 0.75, 1.0\}$, measured through the real discretisation and
projection path (real-image condition; the random-input fan is statistically indistinguishable).
\textit{Top:} every $\alpha > 1/2$ grows geometrically along the scan, faster the larger the value,
while $\alpha \leq 1/2$ does not; dashed lines mark the fp16 ($6.6{\times}10^{4}$) and
bfloat16 ($3.4{\times}10^{38}$) limits. \textit{Bottom:} bounded-regime zoom, in which
$\alpha = 0.25$ decays relative to $1/2$; the step near $d = 128$ is where anti-diagonals leave
the grid boundary.}
\label{fig:supp-alpha}
\end{figure}

\paragraph{Verification at initialisation.}
\Cref{fig:supp-alpha} sweeps $\alpha$ with the model at initialisation, measuring hidden-state
norm growth through the real discretisation and projection path on real image inputs. The exact
rates depend on the image, but the behaviour does not: each value above $1/2$ grows geometrically
along the scan, and the larger the value the faster it grows. At $\alpha = 1$ the state gains tens
of orders of magnitude within a single stage, and even $\alpha = 0.75$ exceeds the bfloat16 range
within the first stage of a $512^2$ input, the regime our high-resolution claims target. At
$\alpha = 1/2$ the norms hold a flat plateau sustained by the input injected at each step, while
$\alpha = 0.25$ decays, with tails $3$--$30\times$ below those of $1/2$.

\section{Influence Geometry: Manhattan vs.\ Raster Distance}
\label{sec:supp-influence}

\begin{figure*}[t]
\centering
\includegraphics[width=\linewidth]{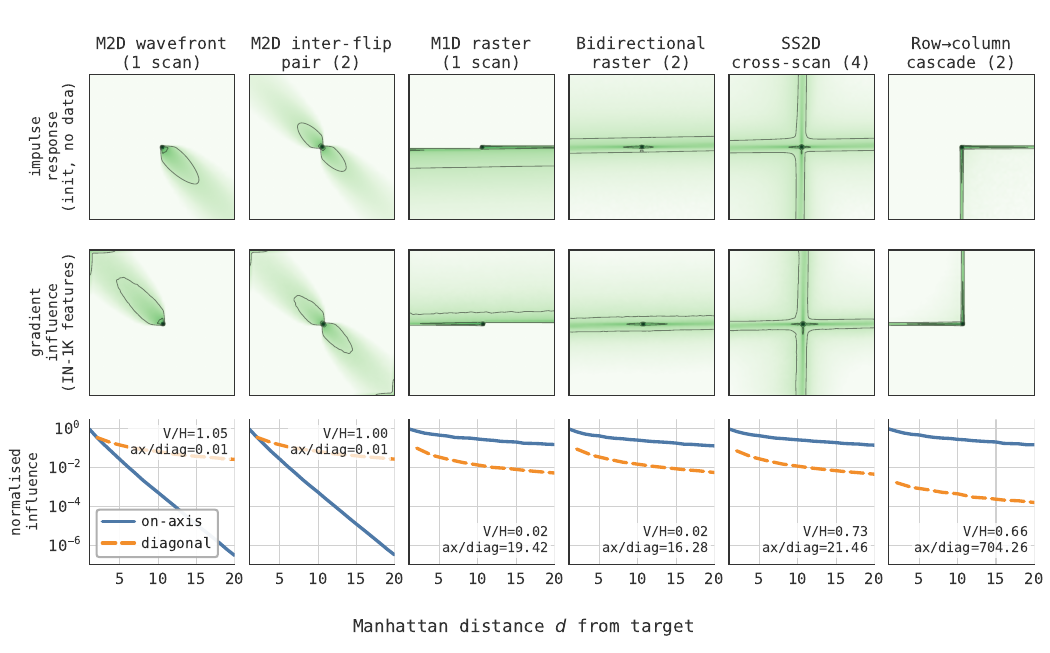}
\caption{Measured influence fields for the M2D wavefront and for alternative scan formulations, with
the local convolution path ablated: impulse response (top), gradient influence
$|\partial \mathbf{y}/\partial \mathbf{x}|$ on ImageNet features (middle), and per-direction decay
(bottom), under matched parameter initialisation in an identical block shell (log colour scale,
per-map normalised, floored at $10^{-6}$). M2D depends on the 2D offset alone, treating the two
axes equally with its mass concentrated toward the diagonal; raster and cascaded scans follow their
flattening order, and SS2D covers the four axes while starving the diagonals.}
\label{fig:supp-influence}
\end{figure*}

\Cref{sec:method} argues that the influence of an input element on an output element is
governed by their 2D offset, rather than by distance along an arbitrary 1D flattening. This
section measures that claim directly: mapping the influence fields of the M2D wavefront and of the
alternative scan formulations shows that only M2D produces a field with this property, with each
1D-flattened alternative failing in its own characteristic way.

For each operator we compute the input-output sensitivity
$|\partial \mathbf{y}_{(t,z)} / \partial \mathbf{x}_{(t',z')}|$, averaged over ImageNet validation
images and over target positions, with the parallel local convolution path ablated so that only the
scan contributes to the measured field. The same probe is applied, under identical conditions and
matched initialisation, to the M2D wavefront and to the four alternative formulations discussed in
the literature: the matched raster scan (M1D), a bidirectional 1D scan, a cascaded row-then-column
scan, and VMamba-style SS2D cross-scanning. \Cref{fig:supp-influence} shows the resulting fields.

M2D's influence is a function of the 2D offset.
No 1D distance function explains the field
($R^2 \leq 0.02$).
An input at Manhattan distance $d$, with $k$ of those steps along one axis, reaches an output through
$\binom{d}{k}$ distinct paths. For example, pixel $(x,y)$ may only influence pixel $(x+3,y)$ through a single route (Right, Right, Right or compactly RRR). Meanwhile, its influence on pixel $(x+2,y+1)$ (which also has a Manhattan distance of 3) propagates along three paths RRD, RDR, DRR.
The influence of each path is weighted by $(1/2)^d$ times the product of the
discretised transition terms along it. These two factors combine to make the field diagonal-dominant. 2D-SSM \cite{baron2DimensionalStateSpace2024} notes the same diagonal bias from this normalisation.
This is confirmed experimentally, with the measured on-axis rate of $0.78$ matching the predicted $0.776$.
Similarly, the classic raster scan's influence is explained well by raster
distance ($R^2$ $0.84$--$0.92$) and poorly by Manhattan distance ($R^2$ $0.42$),
and its field streaks along the flattening direction.
The cascaded row-then-column
scan routes all cross-quadrant influence through a single corner state: its field collapses onto the
two axes, with on-axis influence roughly $700\times$ its diagonal influence at equal Manhattan
distance. SS2D is symmetric across the four
axes, with anisotropy $0.63$--$0.73$, but starves the diagonals by $15$--$21\times$ at equal Manhattan
distance: averaging four 1D scans produces a plus-shaped field, not a neighbourhood.

These distinctions are properties of the scan formulation, not artefacts of initialisation: training
preserves and even sharpens them. For the trained block the anisotropy is $1.21$ for M2D against
$0.07$ for the raster scan, and across the full trained backbone it is $0.95$ against $0.79$. The
trained raster scan's influence is still explained by raster distance ($R^2 = 0.87$).

\begin{figure*}[t]
\centering
\includegraphics[width=\linewidth]{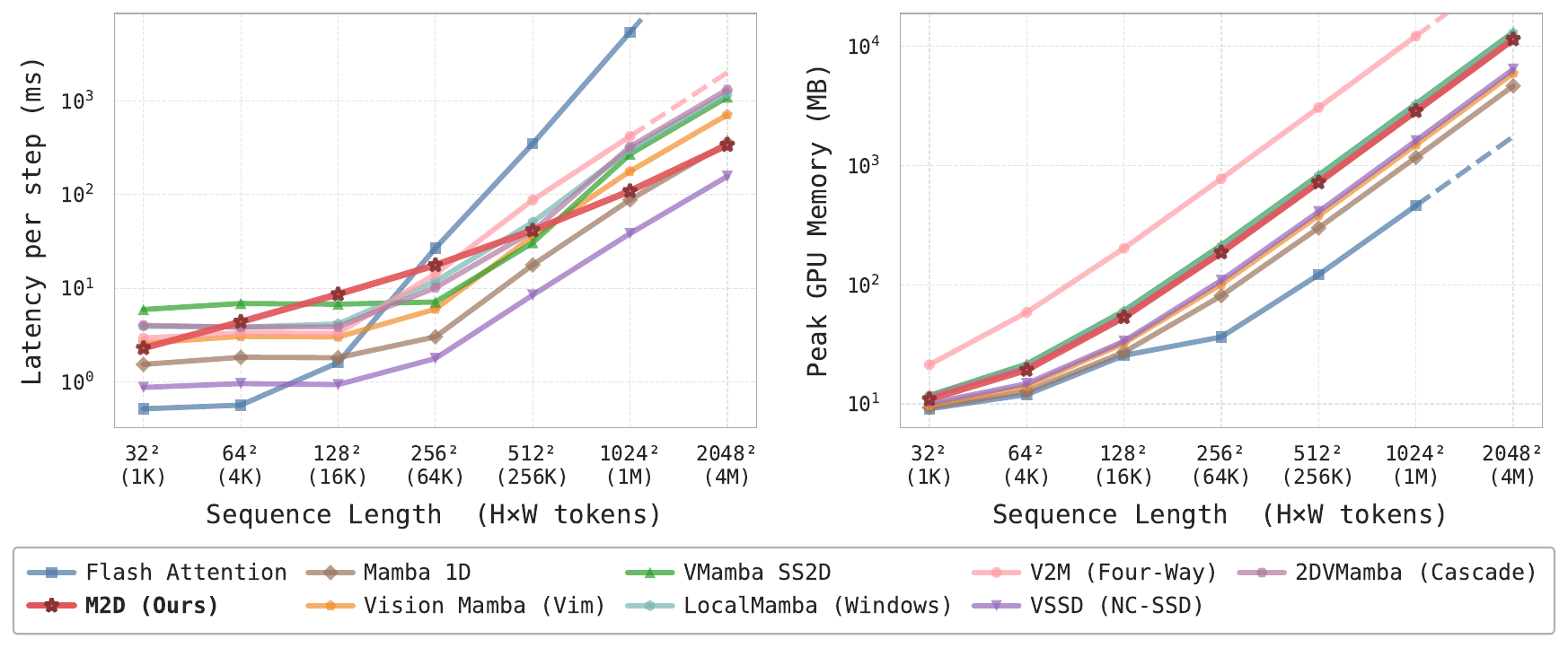}
\caption{Kernel-level scaling of nine token-mixing operators across spatial resolutions ($32^2$--$2048^2$).
\textit{Left:} forward-pass latency (ms, log scale). \textit{Right:} peak GPU memory (MB, log scale).
Dashed lines denote log-log extrapolations: Flash Attention~\cite{flashattn} is capped at $1024^2$ due to intractable runtime; V2M runs out of memory at $2048^2$.}
\label{fig:supp-scan-op-scaling}
\end{figure*}

\section{M1D and SS2D Baseline Construction}
\label{sec:abl-m1d}

The main text reports the scan-dimensionality ablation against our matched 1D baseline, M1D,
with results reproduced in \cref{tab:supp-abl-m1d} for reference, and against VMamba's SS2D
cross-scan~\cite{liuVMambaVisualState2024} in the same shell, tabulated in \cref{tab:abl-ss2d}.
Here we specify the construction of both.

\begin{table}[!htb]
\centering
\caption{Ablation of scanning strategy using a matched M1D baseline on MS-COCO with Mask R-CNN
($1\times$ schedule); $\Delta=$M2D$-$M1D. \textbf{Bold} values indicate M2D outperforms M1D.
Reproduced from the main text.}
\label{tab:supp-abl-m1d}
\resizebox{\linewidth}{!}{%
\scriptsize{\setlength{\tabcolsep}{3pt}
\begin{tabular}{l|rrrr|rrrr}
\toprule
\multirow{2}{*}{Scale} & \multicolumn{4}{c|}{\rule{0pt}{2.15ex}bbox} & \multicolumn{4}{c}{\rule{0pt}{2.15ex}segm} \\
\cline{2-9}
 & \rule{0pt}{2.15ex}mAP & AP$_S$ & AP$_M$ & AP$_L$ & mAP & AP$_S$ & AP$_M$ & AP$_L$ \\
\midrule
N-scale & $\mathbf{+0.1}$ & $\mathbf{+0.5}$ & $\mathbf{+0.1}$ & $\mathbf{+0.4}$ & $0.0$ & $\mathbf{+0.8}$ & $0.0$ & $-0.2$ \\
T-scale &{$ \mathbf{+0.3}$} & $\mathbf{+1.2}$ & $-0.2$ & $\mathbf{+0.8}$ & $\mathbf{+0.3}$ & $\mathbf{+1.0}$ & $-0.2$ & $\mathbf{+0.3}$ \\
\bottomrule
\end{tabular}}}
\end{table}

\begin{table}[!htb]
\centering
\caption{Matched-shell comparison against the SS2D cross-scan baseline on MS-COCO with Mask R-CNN
($1\times$ schedule), N-scale; $\Delta=$M2D$-$SS2D. \textbf{Bold} values indicate M2D outperforms
SS2D.}
\label{tab:abl-ss2d}
\resizebox{\linewidth}{!}{%
\scriptsize{\setlength{\tabcolsep}{3pt}
\begin{tabular}{l|rrrr|rrrr}
\toprule
\multirow{2}{*}{Scale} & \multicolumn{4}{c|}{\rule{0pt}{2.15ex}bbox} & \multicolumn{4}{c}{\rule{0pt}{2.15ex}segm} \\
\cline{2-9}
 & \rule{0pt}{2.15ex}mAP & AP$_S$ & AP$_M$ & AP$_L$ & mAP & AP$_S$ & AP$_M$ & AP$_L$ \\
\midrule
N-scale & $\mathbf{+0.1}$ & $0.0$ & $\mathbf{+0.1}$ & $\mathbf{+0.6}$ & $0.0$ & $-0.2$ & $-0.2$ & $0.0$ \\
\bottomrule
\end{tabular}}}
\end{table}
\begin{table*}[!t]
\centering
\caption{Peak VRAM, forward-only wall time, and forward+backward wall time
relative to the Naive baseline for wavefront scan recomputation strategies.
Benchmarked using a 16-layer operator stack at $64{\times}64$ spatial
resolution on an RTX 5090; relative figures are consistent across spatial
resolutions from $32{\times}32$ to $128{\times}128$. Sequential$^\dagger$
is a correctness reference implemented in Python with no custom CUDA kernel.}
\label{tab:supp-recomp}
\resizebox{0.80\textwidth}{!}{%
{\setlength{\tabcolsep}{9pt}
\begin{tabular}{l|c|c|c|c}
\toprule
Strategy & Peak VRAM (\%) & VRAM Saved (\%) & Fwd Time (\%) & Fwd+Bwd Time (\%) \\
\midrule
Sequential$^\dagger$ & 180              & ---              & ${\sim}9.7{\times}10^3$ & ${\sim}1.1{\times}10^6$ \\
Naive                & 100              & ---              & 100              & 100      \\
\midrule
Partial recomp.      & \textbf{19}      & \textbf{81}      & \textbf{46}      & \textbf{63} \\
Full recomp.         & \textbf{10}      & \textbf{90}      & \textbf{48}      & \textbf{65} \\
\bottomrule
\end{tabular}}}
\end{table*}

Built for exact comparability, the M1D baseline keeps the token-mixer shell of the two
high-resolution stages unchanged, with identical input and output projections and an identical
parallel depthwise-convolution local path. Only the scan operator changes: our joint 2D wavefront
recurrence is swapped for a standard 1D selective scan over the row-major (raster-order) flattening of
the feature map, computed with the vanilla Mamba kernel \cite{guMambaLinearTimeSequence2023}.
As in M2D, each block's scan is unidirectional and the scan direction alternates between
consecutive blocks (top-left-to-bottom-right vs.\ bottom-right-to-top-left), so both models use the same
inter-block strategy for bidirectional context rather than duplicating computation within a block. The
attention stages, block counts, channel widths, drop-path rates and the entire training recipe are
identical, and parameter counts match to within 0.1\% (7.17M vs.\ 7.18M at N-scale; 26.50M vs.\
26.52M at T-scale). The scan formulation is therefore the only varying factor.

The SS2D control is built identically, swapping the same operator for VMamba's cross-scan
implementation. Its four directional scans, with an independent set of SSM parameters ($\mathbf{A}$,
$\mathbf{B}$, $\mathbf{C}$, $\mathbf{D}$, $\Delta$) for each direction, are aggregated within
every block, so it needs no inter-block direction alternation. Total parameter
counts again match to within 0.64\%.

\cref{tab:abl-ss2d} repeats the N-scale comparison against SS2D. The two 2D-aware scans tie,
with overall mAP agreeing to within 0.1, and share a +0.5 bbox AP$_S$ lead over the raster
M1D. The gains are therefore attributable to 2D-aware
scanning itself rather than to our joint formulation specifically. What separates the two is the
cost of that awareness. SS2D reaches parity through four separately parametrised passes, where
M2D uses a single joint recurrence whose markedly better resolution scaling is quantified in
\cref{sec:supp-scan-op}.

\section{Scan Operator Comparison}
\label{sec:supp-scan-op}

We benchmark nine token-mixing operators across spatial resolutions from
$32^2$ (1K tokens) to $2048^2$ (4M tokens). The comparison includes Flash
Attention (quadratic, non-causal), Mamba 1D (unidirectional 1D SSM), Vision
Mamba (Vim, bidirectional), VMamba SS2D (four independent parallel 1D scans),
LocalMamba (window-local 4-direction scan), V2M (two-stage factored
four-direction), VSSD (NC-SSD, non-causal), and 2DVMamba (cascaded
row-then-column scan). All operators are benchmarked on an NVIDIA RTX 5090 in
bfloat16 with batch size 1 and 8 stacked blocks (to reflect realistic
activation memory pressure across a model stage), varying only the spatial
resolution. Results are shown in \cref{fig:scan-op-scaling} and reproduced
here in \cref{fig:supp-scan-op-scaling}.

Flash Attention's quadratic complexity is immediately apparent: each $2\times$
increase in width and height (\ie $4\times$ the number of tokens) multiplies latency by roughly $16\times$ (consistent
with $O(N^2)$), becoming intractable beyond $1024^2$. M2D, by contrast, scales
sub-linearly in latency because the wavefront depth grows as $O(\sqrt{N})$,
yielding an empirical exponent of approximately $0.56$. This gap widens
with resolution: at $1024^2$ M2D is roughly $50\times$ faster than Flash
Attention, and about $2.5\times$ faster than VMamba SS2D and $4\times$ faster
than V2M, both of which pay for multiple independent parallel scans. Although
Mamba 1D is modestly faster at low resolutions, the latency gap closes toward
$2048^2$ as M2D's sub-linear scaling reduces the deficit. V2M runs out of
memory at $2048^2$. The remaining multi-direction operators (VMamba SS2D,
LocalMamba, 2DVMamba) maintain similarly linear scaling but at a consistently
higher latency offset, reflecting the cost of their multiple independent scan
passes.

In terms of peak memory, all SSM-based operators scale linearly. M2D requires
roughly $2.5\times$ more memory than Mamba 1D at high resolution, due to its
2D hidden state, and sits at a similar level to VMamba SS2D. V2M requires
roughly $4\times$ more memory than M2D, consistent with its out-of-memory
failure at $2048^2$. Together, these results show that M2D's wavefront
formulation achieves sub-linear latency scaling of approximately $N^{0.56}$,
against at least $N^{1.0}$ for every multi-directional 1D scan decomposition measured here, at a
constant-factor memory overhead over a single 1D scan.

\section{Kernel Fusion and Recomputation for Memory-Bound SSM Operations}
\label{sec:supp-recomp}

As discussed in the main text, our wavefront scan operator is memory-bound:
the dominant cost is transferring large intermediate tensors through GPU
memory, not arithmetic. This mirrors the 1D selective scan in
Mamba~\cite{guMambaLinearTimeSequence2023}, and makes activation recomputation
particularly effective. Recomputing tensors is arithmetically cheap, yet avoids
the expensive memory traffic of materialising and reloading them between passes.

We benchmark four strategies for kernel fusion and recomputation as detailed below. Each uses a
16-layer stack of isolated scan operators. This configuration is chosen to reflect the cumulative activation
storage across a full model stage. The spatial resolutions tested correspond
approximately to the feature map sizes produced by the convolutional stem and
subsequent downsampling stages, for inputs from $128^2$ to $768^2$:
$32{\times}32$, $64{\times}64$, $128{\times}128$, and $192{\times}192$. All
results in \cref{tab:supp-recomp} are reported relative to the Naive baseline;
the relative savings are consistent across all tested resolutions.

\paragraph{Sequential (reference).}
A pure Python implementation, iterating over spatial positions in anti-diagonal
order using standard PyTorch autograd, with no custom CUDA kernel. Included
solely as a correctness reference; it is roughly two orders of magnitude slower than
the CUDA baseline in the forward pass, and four orders slower for forward and
backward combined, while consuming almost double the memory, as PyTorch retains
the full autograd computation graph.

\paragraph{Naive.}
The CUDA baseline with no recomputation. The backward pass requires three
tensors of shape $(B{\times}H{\times}W{\times}E_d{\times}N)$ to remain resident on the GPU between the forward
and backward passes, where $B$ is the
batch size, $H{\times}W$ the spatial extent, $E_d$ the channel dimension, and
$N$ the SSM state dimension. These tensors are the hidden state sequence $\mathbf{h}$, and the
discretised gate tensors $\overline{\mathbf{A}}_{t,z} = \exp(\Delta \cdot A)$
and $\overline{\mathbf{B}}_{t,z} = \Delta \cdot \mathbf{B} \cdot x$ which
constitute the dominant memory cost at all spatial resolutions.

\paragraph{Partial recomputation.}
The hidden state sequence $\mathbf{h}$ is discarded after the forward pass and
recomputed by re-executing the forward kernel at the start of the backward pass.
The discretised tensors $\overline{\mathbf{A}}_{t,z}$ and
$\overline{\mathbf{B}}_{t,z}$ are similarly never stored as persistent
activations. Instead, the backward kernel recomputes $\exp(\Delta \cdot A)$ on-the-fly
in registers, retaining only the lower-dimensional continuous-time parameters
$(\Delta_t, \Delta_z, \mathbf{B}_t, \mathbf{B}_z) \in
\mathbb{R}^{B{\times}H{\times}W{\times}E_d}$ in the saved context. This
eliminates all three large tensors from the activation budget.

\paragraph{Full recomputation.}
Extends the partial strategy by additionally discarding all projected
activations after the forward pass, retaining only the raw input $x$ and the
learned projection weight matrices. During backpropagation, the input
projections are re-executed to reconstruct the projection computation graph,
after which $\mathbf{h}$ is recomputed as in the partial strategy. This further
reduces the activation footprint at a small additional arithmetic cost.

The Sequential baseline underscores the need for a custom kernel, incurring
${\sim}97{\times}$ the forward wall time, ${\sim}11{,}000{\times}$ the combined
forward and backward wall time, and ${\sim}180\%$ the peak VRAM of the CUDA
baseline. Among the CUDA strategies, recomputation is (as
predicted by the memory-bound characterisation) faster end-to-end, not merely
more memory-efficient. Partial recomputation reduces peak VRAM to
${\sim}19\%$ of Naive while also reducing forward+backward wall time by
${\sim}37\%$ and forward-only wall time by ${\sim}54\%$. The forward speedup
arises because omitting the write of $\mathbf{h}$ and the discretised gate
tensors to global memory is cheaper than materialising them, even without a
backward pass. Full recomputation reduces peak VRAM further to ${\sim}10\%$ of
Naive at a marginal ${\sim}2\%$ Fwd+Bwd overhead relative to Partial, making it
the preferred strategy when VRAM is the primary constraint.

At $192{\times}192$ spatial resolution (corresponding to a $768{\times}768$
input at $4{\times}$ stride), the Naive baseline exceeds the 32\,GB memory
budget. As PyTorch allocates the full \texttt{ctx} tensors ($\mathbf{h}$,
$\overline{\mathbf{A}}_{t,z}$, $\overline{\mathbf{B}}_{t,z}$) during the
forward phase in anticipation of a potential backward, the Naive
variants without recomputation cannot even run inference (forward-only)
at this resolution on a 32\,GB GPU. Both
recomputation variants remain well within budget at $192{\times}192$.
Recomputation is therefore not merely a training optimisation but a prerequisite
for high-resolution inference in downstream tasks.

\section{FLOPs and the Accuracy--Efficiency Trade-off}
\label{sec:supp-flops}

On MS-COCO, M2D-S uses more FLOPs than several baselines in its group ($549$\,G against $325$\,G for
VSSD-S, $348$\,G for ConvNeXt-S, $349$\,G for VMamba-S and $354$\,G for Swin-S). The gap comes from
block distribution, not from the wavefront kernel. M2D-S places 12 blocks in high-resolution stage~2 (depths $[3,12,14,3]$), where
Mask R-CNN's $1280{\times}800$ compute is concentrated, whereas VMamba-S packs its depth into the
downsampled stage~3 (depths $[2,2,27,2]$). That compute is repaid where it is spent, at $+1.5$ to
$+2.0$ box AP over VSSD-S, VMamba-S and V2M-B$^{*}$ at the same scale.

The trade-off inverts on ADE20K, where high-resolution scaling dominates the cost: M2D-T runs at
$485$\,G against $941$\,G for VSSD-T and $948$\,G for VMamba-T, and M2D-S at $768$\,G against
$1028$\,G for both VMamba-S and ConvNeXt-S, with higher mIoU in each case. The exceptions run the
other way: LocalVim-T reaches $43.4$ mIoU at $260$\,G against
M2D-N's $43.1$ at $355$\,G, and V2M-B$^{*}$ is lighter than M2D-S at $655$\,G, though with more
parameters and $50.8$ against $51.7$ mIoU. Overall, M2D trades
FLOPs for accuracy on COCO at S-scale, and sits on the accuracy--FLOPs frontier on
ADE20K at T- and S-scale.

\end{document}